\pdfoutput=1
\documentclass{article}

\PassOptionsToPackage{numbers, sort&compress}{natbib}

\usepackage[main, final]{neurips_2026}

\usepackage[utf8]{inputenc}
\usepackage[T1]{fontenc}
\usepackage{hyperref}
\usepackage{url}
\usepackage{booktabs}
\usepackage{amsfonts}
\usepackage{nicefrac}
\usepackage{microtype}
\usepackage{xcolor}

\usepackage{multirow}
\usepackage{amsmath}
\usepackage{amssymb}
\usepackage{graphicx}
\usepackage{colortbl}
\usepackage{wrapfig}
\usepackage{placeins}
\usepackage{algorithm}
\usepackage{algpseudocodex}
\usepackage{subcaption}
\usepackage{adjustbox}
\usepackage{longtable}
\usepackage{pdflscape}
\usepackage{tabularx}
\usepackage{array}

\definecolor{myframepink}{RGB}{72,138,176}

\hypersetup{
    colorlinks=true,
    allcolors=myframepink
}

\title{Posterior-First Neural PDE Simulation: Inferring \\ What Downstream Tasks Need from a Single Field}

\author{
Wenshuo Wang$^{1}$, Fan Zhang$^{2,3,4}$\thanks{Corresponding author.} \\
$^{1}$ School of Future Technology, South China University of Technology, China \\
$^{2}$ State Key Laboratory of Ocean Sensing, \\
ZJU-Hangzhou Global Scientific and Technological Innovation Center, \\
Zhejiang University, Hangzhou, 311215, China \\
$^{3}$ Ocean College, Zhejiang University, Zhoushan, Zhejiang 316000, China \\
$^{4}$ Kavli Institute for Astrophysics and Space Research, \\
Massachusetts Institute of Technology, Cambridge, MA, 02139, USA \\
\texttt{202364870251@mail.scut.edu.cn}, \texttt{f.zhang@zju.edu.cn}
}

\begin{document}

\maketitle
\begingroup
\renewcommand{\thefootnote}{}
\footnotetext{Our code is available at \url{https://github.com/VinceWangVW/Posterior-First}.}
\endgroup

\begin{abstract}
Neural PDE simulators often receive only a single observed field at deployment. In this setting, a field-to-future predictor can collapse distinct latent problem states into the same deterministic interface, losing the ambiguity needed for reliable rollout and downstream decisions. We propose \textbf{posterior-first neural PDE simulation}, a task-to-target-to-interface paradigm: the declared downstream tasks determine the minimal task-sufficient problem state to infer, and the posterior over that state conditions the rollout and downstream queries. This posterior can be supplied by amortized posterior estimators, including simulation-based inference with learned summaries; our contribution is the target, the interface, and their theory. We show that Bayes downstream values factor through this posterior, proper scoring rules on refinement labels identify it as their population-risk minimizer, and deterministic collapse incurs an ambiguity barrier whenever the true posterior is non-Dirac. Synthetic exact-ambiguity experiments show that point-versus-posterior gaps track the predicted barrier. On metadata-hidden PDEBench tasks, posterior recovery reduces pooled rollout nRMSE from 0.175 to 0.132, closing 59.4\% of the direct-to-oracle gap. Matched controls attribute the gains to task-aligned targeting and posterior deployment, and the gains persist with a latent-diffusion backbone. These results suggest that single-observation neural PDE simulation should be posterior-first rather than monolithic field-to-future prediction.
\end{abstract}

\section{Introduction}

Across science and engineering, many central problems in forecasting, design, and control require fast simulation of systems governed by partial differential equations. Neural PDE simulators have therefore become an important surrogate modeling paradigm, because they can amortize the cost of repeated numerical solves~\cite{li2021fno,lu2021deeponet,lippe2023pderefiner}. We focus on the single-observation regime, where the simulator receives one field and must predict what happens next. This regime matters because a single observed field is often the practical input available at deployment, while the simulator is still expected to support long-horizon prediction and other downstream scientific use. But this regime is fundamentally difficult. One field need not uniquely determine the future, because hidden state, coefficients, operators, boundary conditions, or memory may remain unresolved~\cite{raissi2020hfm,raissi2019pinns,raissi2018dhpm,buitragoruiz2025memory}.

Existing neural PDE simulators differ in what they receive at inference time and how they handle hidden problem-state information before prediction. Some treat simulation as direct prediction from the visible field \cite{li2021fno,lu2021deeponet,lippe2023pderefiner}; others reduce the difficulty with extra problem information or richer context, such as equations, coefficients, boundary conditions, or short trajectories \cite{zhou2025unisolver,liu2024prose,yang2023icon,morel2025disco,serrano2025zebra}; and a related line explicitly recovers hidden state, parameters, operators, or memory before prediction \cite{williams2024shred,raissi2020hfm,raissi2019pinns,buitragoruiz2025memory}. Amortized simulation-based inference (SBI) goes further and learns full posteriors over simulator parameters, often from learned summary statistics of the observation \cite{cranmer2020sbi,radev2022bayesflow,chen2021nss}. Yet single-observation simulation still lacks a principled reusable intermediate object. The missing problem state is usually left implicit, bypassed with extra information, or only partially recovered, and even a full posterior targets a parameter chosen in advance rather than the distinctions that downstream tasks require. As a result, ambiguity unresolved by one field is not preserved for downstream prediction or decision.

We address this gap with \textbf{posterior-first neural PDE simulation}:
\[
x_t \longmapsto q_\phi\left(z_t \mid x_t\right) \longmapsto p_\psi\left(y \mid x_t, z_t\right).
\]
Here \(x_t\) is the observed field at inference time, \(q_\phi\left(z_t \mid x_t\right)\) is a posterior over latent problem states compatible with that field, and \(p_\psi\left(y \mid x_t, z_t\right)\) denotes downstream prediction conditioned on the field and such a state. This factorization makes explicit what single-observation simulation must do under ambiguity: first infer which hidden problem states remain plausible from the visible field, then evolve or answer downstream queries conditional on that uncertainty. Its first stage is an amortized posterior estimator of the kind used in SBI, with fixed or learned summaries. What posterior-first adds is a task-to-target-to-interface construction: the declared tasks determine which hidden distinctions \(z_t\) must retain, and the inferred posterior stays operative in rollout. Concretely, in Section~\ref{sec:formulation} we define the latent problem state and the posterior-first factorization and give a construction of this target from a declared task family. We then show in Section~\ref{sec:theory} that the Bayes-optimal reusable intermediate object is a posterior, that proper scoring rules identify this posterior as their population-risk minimizer, and that deterministic point latents face an explicit ambiguity barrier. To make this formulation concrete, we instantiate in Section~\ref{sec:method} a minimal posterior-first simulator that infers a structured problem-state posterior and conditions evolution on it through a matched rollout backbone and a theorem-facing training objective.

Across the two empirical layers in Section~\ref{sec:empirical}, posterior-first models consistently outperform matched direct-rollout baselines, capacity-matched controls, and literature alternatives. Specifically, on the theorem-facing synthetic benchmark, they close most of the direct-to-oracle gap under controlled ambiguity. On the public PDEBench layer, they improve rollout error on every family, reducing the pooled public \(\mathrm{nRMSE}\) from 0.175 to 0.132 while closing about 60\% of the hidden-metadata oracle gap. Matched controls then separate estimator, target, and interface: task-aligned targeting and posterior-mixture deployment carry the gains under a fixed SBI estimator, posterior sampling beats auxiliary labels and MAP conditioning with a latent-diffusion backbone, a frozen posterior encoder transfers to held-out tasks with few labels, and the advantage persists in a compound stress test. These results support our core claim: single-observation neural PDE simulation should be posterior-first rather than a monolithic field-to-future map. In Section~\ref{sec:discussion} we explain why this matters: making pre-rollout problem-state recovery explicit changes not only how neural simulators are built, but also what scientific surrogate models should be optimized and evaluated for.

\section{Related Work}\label{sec:related_work}

\subsection{Types of Inputs to Neural Simulators}

Research on neural PDE simulation can be classified by the inputs available to the simulator at inference time. A first class uses the observed field itself as the input and learns either a solution operator or a rollout map directly from fields or coordinates. FNO and DeepONet are canonical examples of this regime \cite{li2021fno,lu2021deeponet}. PDE-Refiner is a representative long-horizon rollout model in the same setting \cite{lippe2023pderefiner}. A second class gives the simulator structured problem information in addition to the field. This information can include symbolic equations, coefficients, boundary conditions, geometry, or multimodal operator descriptions. Unisolver and PROSE exemplify this explicit-condition regime \cite{zhou2025unisolver,liu2024prose}. A third class gives the simulator richer context before or during prediction. This context can take the form of short trajectories, contextual examples, or prompts from related dynamics. ICON, DISCO, and Zebra are representative methods in this class \cite{yang2023icon,morel2025disco,serrano2025zebra}.

Taken together, these lines of work still leave single-observation simulation without a principled reusable intermediate object: direct-rollout methods leave the missing problem state implicit, whereas methods that avoid this difficulty do so by supplying extra information from outside the observation itself. Our work addresses this missing object directly with a posterior-first formulation of neural PDE simulation. This matters because it turns unresolved problem-state ambiguity into an explicit reusable simulator interface, rather than leaving it hidden inside end-to-end rollout or bypassing it with extra inputs, and thereby replaces an architecture-level modeling habit with a general principle for how scientific simulators should be built, reasoned about, and evaluated.

\subsection{Problem-State Recovery Methods and What They Recover}

Among methods that recover missing information before prediction, prior work falls into four broad groups: reconstructing hidden field state from partial observations \cite{williams2024shred,raissi2020hfm}, identifying equations, parameters, coefficients, or source terms \cite{raissi2019pinns,raissi2018dhpm}, inferring operator or dynamics representations from richer context \cite{yang2023icon,morel2025disco,serrano2025zebra}, and treating memory as part of the latent state \cite{buitragoruiz2025memory}. Taken together, these methods remain partial: they recover only one slice of the hidden problem state or rely on richer context than a single observed field. As a result, ambiguity that is joint across hidden state, coefficients, operators, boundary conditions, and memory is still not exposed as a reusable object from one field. Our formulation instead targets a posterior over the joint task-sufficient problem state, so that unresolved ambiguity can be carried forward explicitly into downstream evolution and decision-making.

The closest machinery is amortized posterior estimation. Simulation-based inference (SBI) trains neural posterior estimators \(q_\phi(\theta\mid s_\eta(x))\) with a fixed or jointly learned summary \(s_\eta\) of the observation \cite{papamakarios2016fast,cranmer2020sbi,radev2022bayesflow,chen2021nss,tejero2020sbi}, using spline flows, adversarial training, or conditional diffusion \cite{durkan2019nsf,ramesh2022gatsbi,chen2025cdiff}, and PDE pipelines infer a chosen parameter or function and may simulate from posterior samples \cite{hull2024sbi,moss2025fnope}. Supervised metadata conditioning feeds one predicted hidden label to the simulator, and latent diffusion emulators learn future-trajectory uncertainty once the simulation parameters are supplied \cite{rozet2025lola}. These methods provide the estimator and the backbone, but none derives the target from the downstream task family or keeps its posterior operative in a neural rollout: SBI targets a parameter chosen in advance, hard conditioning collapses the posterior to a point, and a direct generator can match one task's future marginal without exposing which hidden problem produced it. Posterior-first simulation uses these components as they are and contributes this task-to-target-to-interface construction. This matters because finite simulation and model capacity are then spent on distinctions that can change the requested outputs, while the remaining ambiguity is carried forward as several plausible rollouts.

\section{Posterior-First Problem-State Formulation}\label{sec:formulation}

We formalize what to infer from one field, why the inferred object is a posterior, and how it is obtained from benchmark labels. The single-observation regime is hard because one observed field need not identify hidden state, coefficients, operators, boundary conditions, or memory relevant to downstream tasks. We therefore separate the observed field from the latent world that generated it: let \(W_t\) be the complete latent world state, and let \(X_t=\mathcal O(W_t)\) be the visible field under observation map \(\mathcal O\). We write \(x_t\) for a realized observation. Downstream scientific uses are indexed by \(\tau\in\mathcal T\), with target variable \(Y_t^\tau\); the family \(\mathcal T\) may include future fields, hidden-state queries, coefficient or operator queries, and uncertainty-sensitive decisions.

The reusable interface should not copy all of \(W_t\). The full latent world is sufficient but may contain distinctions that no downstream task can ever use. We call \(Z_t=g(W_t)\) \emph{task-sufficient} if, once the observation \(X_t\) and \(Z_t\) are known, the remaining details of \(W_t\) do not change the law of any task target:
\begin{equation}
Y_t^\tau \perp W_t \mid (X_t,Z_t),
\qquad \tau\in\mathcal T.
\label{eq:form_sufficient}
\end{equation}
Sufficiency alone is still too weak, since many overly detailed states, including \(W_t\) itself, satisfy \eqref{eq:form_sufficient}. We therefore identify, within each observation fiber, latent worlds that are indistinguishable to the whole task family. Let \(\mathcal L(\cdot\mid\cdot)\) denote conditional law, and define
\begin{equation}
w\sim_{\mathcal T} w'
\iff
\mathcal O(w)=\mathcal O(w')
\ \text{and}\
\mathcal L\!\left(Y_t^\tau\mid W_t=w\right)
=
\mathcal L\!\left(Y_t^\tau\mid W_t=w'\right)
\ \ \text{for all }\tau\in\mathcal T .
\label{eq:form_equiv}
\end{equation}
Let \(Z_t^\star=g^\star(W_t)\) be the quotient label induced by this relation. It is the minimal task-sufficient problem state: given the observed field, it keeps exactly the latent distinctions that can change at least one downstream law and discards the rest. Appendix~\ref{app:form_world} gives the formal quotient construction and proves that every other task-sufficient state is a refinement of \(Z_t^\star\).

Even the minimal state is not necessarily identified by one observed field. Multiple values of \(Z_t^\star\) can remain compatible with the same \(x_t\). The object to infer is therefore not a point latent, but the posterior over task-sufficient problem states:
\begin{equation}
\pi_t(\cdot\mid x_t)
:=
\mathbb P\!\left(Z_t^\star\in\cdot\mid X_t=x_t\right).
\label{eq:form_posterior}
\end{equation}
While \(Z_t^\star\) is minimal for the whole family, a single task depends only on a coarser projection of it. For a task \(\tau\), let \(z_\tau=g_\tau(z)\) be this task-relevant projection of \(z\), which merges states that induce the same law of \(Y_t^\tau\), and let \(\pi_{t,\tau}(\cdot\mid x_t):=(g_\tau)_\#\pi_t(\cdot\mid x_t)\). With \(x_t\) and \(\tau\) held fixed, prediction integrates only the uncertainty in this projection:
\begin{equation}
p_\tau(y\mid x_t)
=
\int p_\tau(y\mid x_t,z_\tau)\,\pi_{t,\tau}(dz_\tau\mid x_t),
\label{eq:form_factorization}
\end{equation}
where \(p_\tau(y\mid x_t,z_\tau)\) denotes the predictive law of \(Y_t^\tau\) given the field and the projected state, written in density notation. Integrating over all of \(Z_t^\star\) gives the same law by construction of \(g_\tau\). Thus the visible field is first converted into uncertainty over plausible problem states, and downstream prediction is then conditional on that uncertainty. Appendix~\ref{app:form_posterior} gives the measure-theoretic derivation.

In benchmarks, \(Z_t^\star\) is the theorem-level object but is not directly annotated. We instead observe a countable semantic refinement \(C_t\), with \(Z_t^\star=h(C_t)\), whose labels record the hidden problem-state slices available for supervision. If \(r_t(c\mid x_t)=\mathbb P(C_t=c\mid X_t=x_t)\) is the refinement posterior, then the desired task-sufficient posterior is its pushforward:
\begin{equation}
\pi_t(z\mid x_t)
=
\sum_{c:\,h(c)=z} r_t(c\mid x_t).
\label{eq:form_pushforward}
\end{equation}
Appendix~\ref{app:form_refinement} formalizes this refinement bridge. Hence the core inference target is \(\pi_t(\cdot\mid x_t)\), represented in experiments by a supervised refinement posterior.

The map \(h\) is fixed by design rather than learned, and it must preserve sufficiency. On a finite catalogue with common support, let \(K_\tau(\cdot\mid x,c):=\mathbb P(Y_t^\tau\in\cdot\mid X_t=x,C_t=c)\), identify \(c\sim_{\mathcal T}c'\) when these kernels agree for every declared task and almost every \(x\), and set \(h(c):=[c]_{\mathcal T}\). This \(h\) satisfies \(Y_t^\tau\perp C_t\mid(X_t,h(C_t))\) and is the coarsest map that does so (Appendix~\ref{app:form_refinement}). Operationally, one declares the task family and uses the governing equations or simulator dependencies, together with the task functionals, to keep every catalogue factor that can change some task kernel, merging only distinctions certified to be task-invariant; keeping an uncertified distinction preserves sufficiency at the price of an over-refined target. In our Shallow-Water protocol, for example, \(x_t\) is the initial water-height field and \(C_t\) indexes eight joint source/boundary states; a boundary query depends only on the boundary coordinate, whereas rollout may depend on both. Single-observation neural PDE simulation should be posterior-first: problem-state posterior inference followed by conditional evolution, rather than a monolithic field-to-future map.

\section{Why the Reusable Intermediate Object Is a Posterior}
\label{sec:theory}

Section~\ref{sec:formulation} defined the posterior over the minimal task-sufficient problem state. We now argue that it is the right object between observation and evolution because it is sufficient, learnable, irreplaceable by a point, and worth learning well: Bayes decisions depend on the observation only through it (Section~\ref{sec:theory_object}); proper scoring rules on refinement labels identify it, whereas any point latent remains an ambiguity barrier away from it (Section~\ref{sec:theory_learnable}); and its error bounds the downstream utility gap, while rollout-only training does not identify it (Section~\ref{sec:theory_utility}). Appendix~\ref{app:posterior_theory} contains the full derivations.

\subsection{The Correct Reusable Intermediate Object Is a Task-Sufficient Posterior}
\label{sec:theory_object}

Fix a task \(\tau\in\mathcal T\), action space \(\mathcal A_\tau\), action \(a\in\mathcal A_\tau\), and task loss \(\ell_\tau(a,Y_t^\tau)\). Let \(\rho_\tau(a,z):=\mathbb E[\ell_\tau(a,Y_t^\tau)\mid Z_t^\star=z]\). For a realized observation \(x_t\), the conditional Bayes value satisfies
\begin{align}
\mathcal V_\tau(a,x_t)
&:=
\mathbb E\!\left[\ell_\tau(a,Y_t^\tau)\mid X_t=x_t\right] \notag\\
&=
\mathbb E\!\left[\rho_\tau(a,Z_t^\star)\mid X_t=x_t\right] \notag\\
&=
\int \rho_\tau(a,z)\,\pi_t(dz\mid x_t).
\label{eq:theory_value_factorized}
\end{align}
Task-sufficiency gives the second line, and conditioning on \(\pi_t(\cdot\mid x_t)\) gives the third. Hence Bayes actions depend on \(x_t\) only through this posterior, making the reusable object posterior-valued rather than point-valued. For the benchmark refinement, the same value is obtained from \(r_t\) via the pushforward in \eqref{eq:form_pushforward}.

\subsection{The Posterior Is Learnable and Point Latents Are Ambiguity-Limited}
\label{sec:theory_learnable}

Let \(\mathcal C=\mathrm{range}(C_t)\), let \(r_X(\cdot):=r_t(\cdot\mid X_t)\) be the true refinement posterior, and let \(q_X(\cdot)\) be a predicted distribution on \(\mathcal C\). For the log score \(\mathcal L_{\log}(q):=\mathbb E[-\log q_X(C_t)]\), the excess risk over the Bayes risk \(\mathcal L_{\log}^\star\) is
\begin{equation}
\mathcal L_{\log}(q)-\mathcal L_{\log}^\star
=
\mathbb E\!\left[
\sum_{c\in\mathcal C} r_X(c)\log\frac{r_X(c)}{q_X(c)}
\right]
=
\mathbb E\!\left[\mathrm{KL}(r_X\,\|\,q_X)\right].
\label{eq:theory_log}
\end{equation}
The log score therefore already identifies the refinement posterior: for a well-specified model class, its unique population-risk minimizer is \(q_X=r_X\). The multiclass Brier score \(\mathcal L_{\mathrm{Br}}(q):=\mathbb E[\|q_X-e_{C_t}\|_2^2]\) plays a different role; let \(e_c\) be the one-hot distribution at label \(c\). Since \(\mathbb E[e_{C_t}\mid X_t]=r_X\), conditioning on \(X_t\) gives the orthogonal decomposition
\[
\mathbb E\!\left[\|q_X-e_{C_t}\|_2^2\mid X_t\right]
=
\|q_X-r_X\|_2^2
+
\mathbb E\!\left[\|r_X-e_{C_t}\|_2^2\mid X_t\right].
\]
Thus, with \(\mathcal L_{\mathrm{Br}}^\star\) denoting the Bayes Brier risk,
\begin{equation}
\mathcal L_{\mathrm{Br}}(q)-\mathcal L_{\mathrm{Br}}^\star
=
\mathbb E\!\left[\|q_X-r_X\|_2^2\right].
\label{eq:theory_brier}
\end{equation}
This exact squared-\(L_2\) regret is not needed for identification; it supplies the geometry of point collapse below and the finite-\(\mathcal C\) utility bound in Section~\ref{sec:theory_utility}.

The same calculation explains why deterministic point latents are insufficient in genuinely ambiguous observations. A point predictor \(\hat c(X_t)\) represents a one-hot distribution. For any posterior \(r\), \(\|e_c-r\|_2^2=1-2r(c)+\|r\|_2^2\), so the best one-hot approximation incurs
\begin{equation}
\Gamma_{\mathrm{amb}}(r)
:=
\min_c\|e_c-r\|_2^2
=
1+\|r\|_2^2-2\|r\|_\infty
\ge
(1-\|r\|_\infty)^2.
\label{eq:theory_gamma}
\end{equation}
This ambiguity penalty vanishes only for Dirac posteriors. Therefore, when one field leaves multiple problem states plausible, a point latent is separated from the posterior target by an irreducible statistical gap.

\subsection{Posterior Quality Controls Simulator Utility}
\label{sec:theory_utility}

Posterior recovery matters because downstream value is linear in the posterior. For an approximate refinement posterior \(q_{x_t}\), define \(\mathcal V_\tau^q(a,x_t):=\sum_{c\in\mathcal C}\rho_\tau(a,h(c))q_{x_t}(c)\). If \(B_\tau:=\sup_{a\in\mathcal A_\tau,c\in\mathcal C}|\rho_\tau(a,h(c))|<\infty\), then for every action \(a\),
\begin{equation}
\big|
\mathcal V_\tau^q(a,x_t)-\mathcal V_\tau(a,x_t)
\big|
\le
B_\tau\,
\|q_{x_t}-r_t(\cdot\mid x_t)\|_1.
\label{eq:theory_value_err}
\end{equation}
Let \(a_\tau^q(x_t)\) minimize the approximate value \(\mathcal V_\tau^q\), and let \(a_\tau^\star(x_t)\) minimize the true value \(\mathcal V_\tau\). Then
\begin{equation}
\mathcal V_\tau\!\big(a_\tau^q(x_t),x_t\big)
-
\mathcal V_\tau\!\big(a_\tau^\star(x_t),x_t\big)
\le
2B_\tau\,
\|q_{x_t}-r_t(\cdot\mid x_t)\|_1.
\label{eq:theory_task_excess}
\end{equation}
Proper scoring losses then translate into utility control. By Pinsker's inequality and \eqref{eq:theory_log}, and, if \(\mathcal C\) is finite, by \(\|v\|_1\le\sqrt{|\mathcal C|}\|v\|_2\) and \eqref{eq:theory_brier},
\begin{equation}
\mathbb E\!\left[\|q_X-r_X\|_1\right]
\le
\sqrt{2\big(\mathcal L_{\log}(q)-\mathcal L_{\log}^\star\big)},
\qquad
\mathbb E\!\left[\|q_X-r_X\|_1\right]
\le
\sqrt{|\mathcal C|\big(\mathcal L_{\mathrm{Br}}(q)-\mathcal L_{\mathrm{Br}}^\star\big)}.
\label{eq:theory_l1_log}
\end{equation}
Thus posterior metrics are not merely auxiliary diagnostics: they upper-bound the downstream utility gap of the simulator induced by the learned posterior.

Finally, rollout-only hidden states are not generally identified as reusable objects. Suppose training uses only a restricted rollout family \(\mathcal T_0\subsetneq\mathcal T\). If two distinct posteriors \(\pi\neq\pi'\) over \(Z_t^\star\) satisfy \(\int \rho_\tau(a,z)\,\pi(dz)=\int \rho_\tau(a,z)\,\pi'(dz)\) for all \(\tau\in\mathcal T_0\) and \(a\in\mathcal A_\tau\), then every objective built only from \(\mathcal T_0\) is indifferent between them, even though an omitted task in \(\mathcal T\) may distinguish them. A rollout hidden state can therefore be sufficient for the training rollout loss while failing as a reusable interface. Explicit posterior recovery avoids this non-identification by targeting the task-sufficient posterior itself.

\section{A Minimal Posterior-First Simulator}\label{sec:method}

We instantiate the theory with the smallest architectural change that makes the inferred posterior explicit. In experiments, \(z_t\in\mathcal Z\) is the realized label of the finite annotated semantic refinement \(C_t\) used for supervision: its coordinates cover the hidden field state, coefficient or operator regime, boundary class, or memory class, while \(Z_t^\star\) remains the theorem-level quotient and \(x_t\) remains the visible rollout input. Given the observed field \(x_t\), a posterior encoder predicts
\begin{equation}
q_\phi(z\mid x_t)
\approx
\mathbb P\!\left(C_t=z\mid X_t=x_t\right),
\qquad z\in\mathcal Z .
\label{eq:method_posterior}
\end{equation}
The output is a full joint posterior. This countable annotated refinement is the smallest setting in which the proper-scoring identities of Section~\ref{sec:theory} can be tested directly. The encoder is a seven-layer CNN (widths 32 to 256, GroupNorm and SiLU) with a softmax head over the 16/16/8/16 catalogue states of DR/DS/SW/INS or the 64 synthetic states; continuous metadata cells are fitted on the training split and then frozen.

The rollout still uses \(x_t\) as the visible initial condition; the posterior supplies the unresolved problem-state information. Ideally, prediction follows the finite mixture
\[
\sum_{z\in\mathcal Z} q_\phi(z\mid x_t)p_\psi(y\mid x_t,z),
\]
where \(p_\psi\) is a conditional rollout predictor. In the main experiments, we keep one matched rollout backbone and expose the posterior through the embedding summary
\begin{equation}
s_\phi(x_t)
=
\sum_{z\in\mathcal Z}
q_\phi(z\mid x_t)e(z),
\qquad
\hat y
=
F_\psi(x_t,s_\phi(x_t)).
\label{eq:method_summary}
\end{equation}
Here \(e(z)=E_{z,:}\) is a row of a trainable codebook \(E\in\mathbb R^{|\mathcal Z|\times 64}\), so \(s_\phi(x_t)=q_\phi(\cdot\mid x_t)^\top E\) averages, in the Shallow-Water example, the codes of the eight source/boundary states by their posterior weights; a linear projection adds it to the features after each backbone's input embedding (FNO lifting layer, U-Net stem, ConvLSTM encoder, or Transformer patch embedding), it is computed once from \(x_t\) and reused at every rollout step, and \(F_\psi\) has the same rollout family as the direct baseline. Because nonlinear \(F_\psi\) need not commute with posterior averaging, \eqref{eq:method_summary} is a tested low-cost approximation to the mixture; Appendix~\ref{app:method_summary_gap} compares it with explicit mixtures and posterior sampling, which we use when the full predictive law is needed (Section~\ref{sec:controls}).

Training is a single end-to-end optimization that learns the posterior with a proper scoring rule and trains the simulator with the same downstream losses used by the direct baselines:
\begin{equation}
\min_{\phi,\psi}\;
\lambda_{\mathrm{post}}
\mathbb E\!\left[
S\!\left(q_\phi(\cdot\mid x_t),z_t\right)
\right]
+
\sum_{\tau\in\mathcal T_{\mathrm{train}}}
\lambda_\tau
\mathbb E\!\left[
\ell_\tau\!\left(F_\psi(x_t,s_\phi(x_t)),Y_t^\tau\right)
\right].
\label{eq:method_objective}
\end{equation}
Here \(S\) is the log score on \(z_t\), the task losses are those of the direct baselines (normalized rollout MSE on the public benchmark), and the posterior and total task terms have unit weights. The log score trains the encoder directly, and the task losses backpropagate through \(s_\phi\) into the encoder, codebook, projection, and simulator without stop-gradient. The theory guarantees reuse of the object \((x_t,\pi_t)\); because the task losses also shape \(q_\phi\) and \(E\), reuse of the trained encoder is tested empirically (Section~\ref{sec:controls}), while \(F_\psi\) is task-specific. Implementation details and summary-fidelity checks are in Appendix~\ref{app:method_detail}.

The simulator is minimal for the empirical test: posterior-first and direct models share the rollout family, optimization budget, and task supervision. Removing \(s_\phi(x_t)\) gives the field-to-future baseline, and replacing \(q_\phi(\cdot\mid x_t)\) by a point estimate gives the point-latent control. The comparison is designed to isolate latent-state posterior recovery.

\section{Empirical Evaluation}\label{sec:empirical}

Our experiments test three claims. First, when a single field is ambiguous, posterior-valued problem-state inference should outperform point-valued or direct interfaces and approach the oracle state interface. Second, when hidden problem metadata are withheld in public PDEBench tasks, the same posterior-first mechanism should improve standard rollout under matched backbones~\citep{takamoto2022pdebench}. Third, these gains should come from the task-aligned target and the posterior interface rather than from the posterior estimator, hidden-state labels, or the rollout backbone.

\subsection{Experimental setup}\label{sec:exp_setup}

\subsubsection{Datasets and scope}\label{sec:datasets}

We use two benchmark layers. The synthetic layer is an in-house exact-collision generator built from Gray--Scott-style reaction--diffusion rendering~\citep{gray1984autocatalytic,pearson1993complex}. Each example contains one scalar field \(x_t\in\mathbb R^{1\times64\times64}\), a discrete semantic state \(z_t\in\mathcal Z_{\mathrm{syn}}\), the exact posterior \(r_X(\cdot)=\mathbb P(z_t\in\cdot\mid X_t)\), and task targets for future fields, operator class, coefficient value, and uncertainty. We use \(|\mathcal Z_{\mathrm{syn}}|=64\) states and four ambiguity regimes---near-Dirac, mid, high, and very high---formed by increasing the number of compatible semantic states in each collision class. Full generator rules and splits are given in Appendix~\ref{app:emp_synth_data}.

The public layer is a metadata-hidden protocol built on PDEBench~\citep{takamoto2022pdebench} with four forward families, Diffusion-Reaction (DR), Diffusion-Sorption (DS), Shallow-Water (SW), and Incompressible Navier--Stokes (INS), each rolled out for 100 steps from one initial field. Each item exposes only this field \(x_t\), while selected metadata are withheld and used as \(z_t^{\mathrm{pub}}\): regime/coefficient in DR, coefficients in DS, joint source and boundary class in SW, and forcing or viscosity in INS. Direct baselines use only \(x_t\), posterior-first models infer \(q(z_t^{\mathrm{pub}}\mid x_t)\), and oracle controls receive the metadata. Grids, sequence counts, taxonomies, splits, and probe definitions are reported in Appendix~\ref{app:emp_public_data}.

\subsubsection{Compared methods, backbones, and regimes}

We organize the compared methods by the alternative explanation they test. In the synthetic layer, where the true posterior and oracle state are known, we use the full matched control ladder. \texttt{Direct-Point} is the standard deterministic field-to-future rollout model. \texttt{Direct-Dist} keeps the same direct interface but predicts a heteroscedastic output distribution, testing output uncertainty alone. \texttt{Direct-MTL} adds auxiliary hidden-state supervision to the direct model, testing latent supervision without a posterior interface. \texttt{Direct-Large} widens and deepens the direct model, testing capacity. \texttt{Point-$z$} first predicts a single semantic state \(\hat z_t\) and then conditions rollout on it, while \texttt{Posterior-$z$} predicts the full \(q(z_t\mid x_t)\) before rollout. \texttt{RichContext-Posterior-$z$} receives a 4-frame context that resolves most collisions, and \texttt{Oracle-$z$} receives the true latent problem state; these two play the role of the richer-context and explicit-condition classes of Section~\ref{sec:related_work} and serve as upper references, because they receive information that a single field does not provide. Architectures, losses, and configurations are given in Appendix~\ref{app:method_detail} and the released code.

In the public PDEBench layer, the main text keeps the comparison smaller and focused on external validity: \texttt{Direct-Point}, \texttt{Direct-Dist}, \texttt{Posterior-$z$}, and \texttt{Oracle-$z$}, asking whether explicit hidden-metadata posterior recovery closes the direct-to-oracle rollout gap under matched conditions. Appendices~\ref{app:emp_public} and~\ref{app:emp_public_additional} add external probabilistic baselines where their native assumptions fit the metadata-hidden task, namely Approximate Bayesian Neural Operators (ABNO)~\cite{magnani2025abno}, the invertible neural-operator posterior of Kaltenbach et al.~\cite{kaltenbach2023sino}, and the label-free \texttt{DeepEns-Direct}~\cite{lakshminarayanan2017deepens} and \texttt{GateMoE-Direct}~\cite{shazeer2017moe}, to test whether generic probabilistic direct modeling can replace explicit problem-state posterior recovery.

For rollout backbones, we use four common neural-simulator families with complementary inductive biases: U-Net~\cite{ronneberger2015unet}, FNO~\cite{li2021fno}, ConvLSTM~\cite{xingjian2015convlstm}, and Transformer~\cite{vaswani2017attention}. Within each backbone family, all matched internal controls share the same rollout stem, optimization budget, train/validation/test split, and observation protocol. The public experiments pool over the four families and all matched backbones; external probabilistic baselines are evaluated family-wise where their assumptions apply.

\subsubsection{Metrics and evaluation protocol}\label{sec:procedure&metrics}

For the synthetic benchmark, each method is trained on the synthetic training split, selected by validation task-bundle risk, and evaluated on held-out test examples from the same ambiguity regimes. The mechanism metric is the average ambiguity barrier \(\overline{\Gamma}_{\mathrm{amb}}\), the test mean of \(\Gamma_{\mathrm{amb}}(r_i)\) in \eqref{eq:theory_gamma} under the exact posterior \(r_i\); it is the best Brier error that any deterministic point latent can achieve, and we compare it with the observed \emph{Point gap}, the excess latent-state Brier error of \texttt{Point-$z$} over \texttt{Posterior-$z$}. Downstream, we report the remaining direct-to-oracle gap \(G_{\mathrm{rem}}(m):=(R_{\mathrm{full}}(m)-R_{\mathrm{full}}^{\mathrm{ora}})/(R_{\mathrm{full}}^{\mathrm{dir}}-R_{\mathrm{full}}^{\mathrm{ora}})\) of the full task-bundle risk, which is 1 for the direct baseline and 0 for the oracle, and the negative log-likelihood of the 8-step future field (Future NLL).

For the public PDEBench benchmark, each method is evaluated by closed-loop rollout on held-out trajectories with the trajectory-wise relative error nRMSE and the oracle-gap closure \(\mathrm{GapClose}(m):=(\mathrm{nRMSE}^{\mathrm{dir}}-\mathrm{nRMSE}(m))/(\mathrm{nRMSE}^{\mathrm{dir}}-\mathrm{nRMSE}^{\mathrm{ora}})\), which is 0 without improvement over direct prediction and 1 when the oracle gap is fully closed. For stratified analysis, we compute GapClose within four bins of a held-out single-field ambiguity score \(a(x_t)\), with the high-ambiguity value taken from the highest-score bin. To check whether rollout gains coincide with hidden problem-state recovery, we report the strongest metadata probe for each family: accuracy for discrete metadata and \(R^2\) for continuous metadata, averaged across families in the pooled row. The controls in Sections~\ref{sec:controls}--\ref{sec:stress} also use the CRPS of a scalar parameter posterior and the energy score of sampled rollouts, two proper scores for which lower is better. Results are averaged over five training seeds; full definitions are given in Appendix~\ref{app:emp_metrics}.

\subsection{Main results}

\subsubsection{Synthetic exact-ambiguity benchmark: mechanism and downstream closure}

The synthetic benchmark is the only setting in which the exact posterior and the ambiguity barrier are known. Table~\ref{tab:main_synth_alignment} therefore reports the mechanism check predicted by Section~\ref{sec:theory_object}, and Table~\ref{tab:main_synth_downstream} reports the corresponding downstream consequence on the full task bundle.

\begin{table*}[t]
\centering
\begin{minipage}[t]{0.59\textwidth}
    \centering
    \small
    \caption{Synthetic exact-ambiguity mechanism check across ambiguity regimes and identifiability controls.}
    \label{tab:main_synth_alignment}
    \resizebox{\linewidth}{!}{%
    \begin{tabular}{lccc}
    \toprule
    \textbf{Regime / control} & \textbf{\(\mathbb E[\Gamma_{\mathrm{amb}}(r_X)]\)} & \textbf{Point gap} & \textbf{gap / \(\mathbb E[\Gamma_{\mathrm{amb}}]\)} \\
    \midrule
    Near-Dirac control & 0.004 & 0.004 & 1.00 \\
    Mid ambiguity & 0.018 & 0.016 & 0.89 \\
    High ambiguity & 0.031 & 0.028 & 0.90 \\
    Very high ambiguity & 0.047 & 0.042 & 0.89 \\
    \shortstack[l]{4-frame richer-context\\control} & 0.005 & 0.005 & 1.00 \\
    \shortstack[l]{Operator-token Dirac\\control} & 0.003 & 0.003 & 1.00 \\
    \bottomrule
    \end{tabular}}
\end{minipage}
\hfill
\begin{minipage}[t]{0.35\textwidth}
    \centering
    \small
    \caption{Synthetic downstream performance on the full task bundle.}
    \label{tab:main_synth_downstream}
    \resizebox{\linewidth}{!}{%
    \begin{tabular}{lcc}
    \toprule
    \textbf{Method} & \textbf{Future NLL} & \textbf{\(G_{\mathrm{rem}}\)} \\
    \midrule
    \texttt{Direct-Point} & 1.48 & 1.000 \\
    \texttt{Direct-Dist} & 1.22 & 0.732 \\
    \texttt{Direct-MTL} & 1.24 & 0.498 \\
    \texttt{Direct-Large} & 1.20 & 0.598 \\
    \texttt{Point-$z$} & 1.18 & 0.460 \\
    \texttt{Posterior-$z$} & 1.02 & 0.381 \\
    \shortstack[l]{\texttt{RichContext-}\\\texttt{Posterior-$z$}} & 0.92 & 0.205 \\
    \texttt{Oracle-$z$} & 0.81 & 0.000 \\
    \bottomrule
    \end{tabular}}
\end{minipage}
\end{table*}

Table~\ref{tab:main_synth_alignment} shows that the deterministic point-versus-posterior gap is governed by the ambiguity barrier rather than by generic model capacity. As ambiguity increases from the mid to the very-high regime, the expected barrier rises from 0.018 to 0.047 and the observed point gap rises in step from 0.016 to 0.042, while the ratio between them stays concentrated at 0.89--0.90. In the near-Dirac, 4-frame, and operator-token controls, both the barrier and the observed gap collapse to 0.003--0.005. The synthetic mechanism result therefore matches the theorem in both directions: the point-latent penalty appears when a single field is ambiguous, and it disappears again when the observation becomes identifying.

Table~\ref{tab:main_synth_downstream} shows how that mechanism translates into simulator utility. Relative to \texttt{Direct-Point}, a larger direct model and auxiliary hidden-target supervision reduce the remaining direct-to-oracle gap to 0.598 and 0.498, respectively, but neither matches an explicit posterior stage. The decisive comparison is between \texttt{Point-$z$} and \texttt{Posterior-$z$}: replacing a point latent by a posterior lowers future NLL from 1.18 to 1.02 and reduces \(G_{\mathrm{rem}}\) from 0.460 to 0.381. The richer-context and oracle controls give the expected ceilings at 0.205 and 0.000. Appendix~\ref{app:emp_synth} shows that the same ordering persists on the fuller task bundle and across ambiguity regimes (Tables~\ref{tab:app_synth_full}--\ref{tab:app_synth_stratified}), and its rollout-only ablation (Table~\ref{tab:app_synth_rollout_only}) shows why rollout accuracy cannot identify the reusable state: across five training objectives the training-task nRMSE stays at 0.028--0.030, while the hidden-task score rises from 0.61 for rollout-only training to 0.84 with explicit posterior recovery over the full task family. Appendix~\ref{app:method_summary_gap} then checks the implementation approximation directly: replacing the summary interface by an explicit mixture changes synthetic \(G_{\mathrm{rem}}\) only from 0.381 to 0.365, while a four-sample approximation gives 0.373.

Taken together, the synthetic layer supports our theorem-facing claim: when ambiguity can be controlled exactly, explicit posterior inference is the mechanism that explains the downstream gain.

\subsubsection{Public PDEBench metadata-hidden oracle-gap probe}
\label{sec:public_probe}

\begin{wrapfigure}{r}{0.40\linewidth}
    \vspace{-0.9em}
    \centering
    \includegraphics[width=\linewidth]{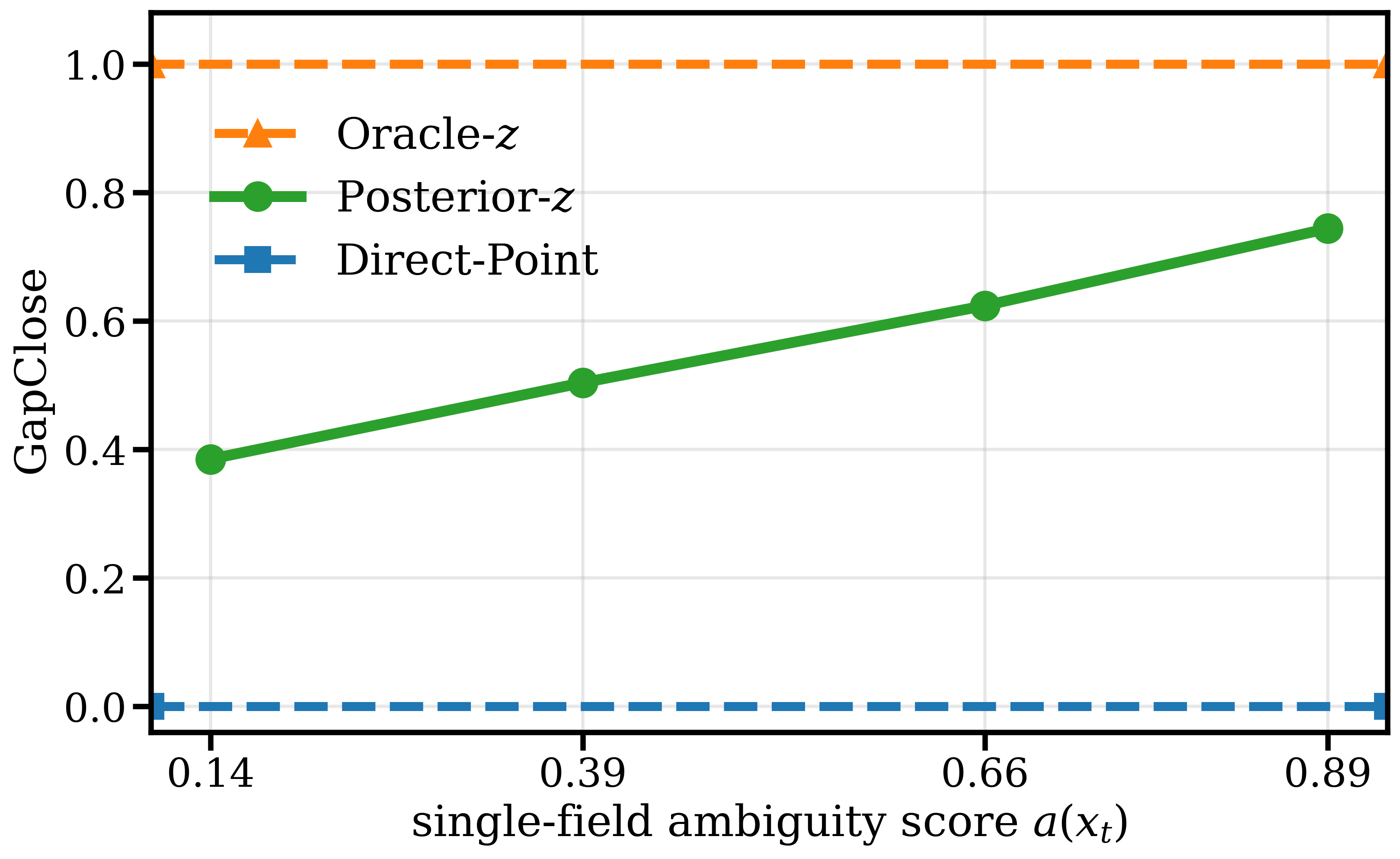}
    \caption{\(\mathrm{GapClose}\) versus held-out single-field ambiguity score \(a(x_t)\) on metadata-hidden PDEBench. Points show \texttt{Posterior-$z$} bin means; dashed lines mark \texttt{Direct-Point} at 0 and \texttt{Oracle-$z$} at 1.}
    \label{fig:main_public_protocol}
\end{wrapfigure}

Because the public benchmark has hidden metadata but no exact posterior labels, this subsection uses three complementary views: gap closure as a function of ambiguity (Figure~\ref{fig:main_public_protocol}), rollout error per family (Table~\ref{tab:main_public_rollout}), and recovery of the hidden metadata (Table~\ref{tab:main_public_probe}). Figure~\ref{fig:main_public_protocol} bins test samples, pooled across families, by a held-out single-field ambiguity score \(a(x_t)\), which estimates how much oracle benefit remains under a single visible field (Appendix~\ref{app:emp_public_data}). As \(a(x_t)\) rises from 0.14 to 0.89 across bins, \texttt{Posterior-$z$} closes more of the direct-to-oracle gap, increasing from 0.385 to 0.744. This gives the ambiguity-stage conclusion: posterior-first simulation helps most on public samples where more hidden problem state remains unresolved.

\begin{table*}[t]
    \centering
    \small
    \caption{Public PDEBench rollout under the metadata-hidden protocol: nRMSE of each method (lower is better) and \(\mathrm{GapClose}\) of \texttt{Posterior-$z$} (higher is better); High-amb.\ uses each family's highest ambiguity bin.}
    \label{tab:main_public_rollout}
    \begin{tabular}{lcccccc}
    \toprule
    \textbf{Family} & \textbf{Direct} & \textbf{Direct-Dist} & \textbf{Posterior-\(z\)} & \textbf{Oracle-\(z\)} & \textbf{\(\mathrm{GapClose}\)} & \textbf{High-amb.\ \(\mathrm{GapClose}\)} \\
    \midrule
    DR & 0.2228 & 0.1957 & 0.1690 & 0.1290 & 0.573 & 0.669 \\
    DS & 0.0285 & 0.0222 & 0.0182 & 0.0125 & 0.641 & 0.740 \\
    SW & 0.0772 & 0.0607 & 0.0532 & 0.0387 & 0.623 & 0.722 \\
    INS & 0.3702 & 0.3311 & 0.2872 & 0.2305 & 0.594 & 0.688 \\
    All & 0.1747 & 0.1524 & 0.1319 & 0.1027 & 0.594 & 0.689 \\
    \bottomrule
    \end{tabular}
\end{table*}

Table~\ref{tab:main_public_rollout} reports the rollout errors and direct-to-oracle gap closure under the metadata-hidden protocol. \texttt{Posterior-$z$} lowers nRMSE in all four public families. In the \texttt{All} row, averaged over the public evaluation settings, nRMSE drops from 0.1747 to 0.1319, compared with 0.1524 for \texttt{Direct-Dist}, which models output uncertainty alone, and the oracle reference 0.1027. This corresponds to 59.4\% direct-to-oracle gap closure overall and 68.9\% on each family's highest-ambiguity bin, whereas the 0.744 of Figure~\ref{fig:main_public_protocol} comes from bins pooled across families. Family-wise closures are all positive, ranging from 0.573 to 0.641, so the gain is not driven by one PDE family. Appendix~\ref{app:emp_public} shows the same ordering for every backbone and against external probabilistic baselines. This gives the rollout-stage conclusion: the ambiguity-sensitive mechanism translates into robust prediction improvement.

Table~\ref{tab:main_public_probe} in Appendix~\ref{app:emp_public} reports the strongest available hidden-metadata probe for each public family. In the \texttt{All} row, the mean probe score rises from 0.628 for the direct baseline to 0.800 for \texttt{Posterior-$z$}, approaching the oracle upper bound of 0.901. The improvement appears in every family, for both discrete metadata accuracy and continuous-metadata \(R^2\). That appendix also reports calibration diagnostics, and Appendix~\ref{app:method_summary_gap} checks the summary-interface approximation. This gives the hidden-state conclusion: the rollout gain comes with better recovery of withheld problem-state information, not only better marginal predictive uncertainty.

Taken together, the public layer supports the practical half of the claim: posterior-first simulation improves standard PDEBench rollout by recovering hidden problem-state information that direct single-field predictors leave implicit.

\subsubsection{Estimator, target, and interface controls}
\label{sec:controls}

Because any amortized posterior estimator can implement the posterior stage, we test whether the gains come from the estimator or from the target and interface (Appendix~\ref{app:emp_controls} gives protocols and five-seed standard errors). In Test~A, on PDEArena's Navier--Stokes task with hidden buoyancy~\cite{gupta2023pdearena}, a learned CNN summary improves posterior CRPS and final-quarter rollout Energy over dimension-matched PCA (from 0.2116/0.1517 to 0.1916/0.1459), whereas a spline-flow NPE~\cite{durkan2019nsf} and a conditional-diffusion NPE~\cite{chen2025cdiff} on the same summary are comparable. In Test~B, the synthetic benchmark gains a task-irrelevant eight-state nuisance \(U\), so \(C=(Z^\star,U)\) and \(h(z,u)=z\). With the NPE family fixed, inferring \(Z^\star\) directly instead of \(C\) followed by exact pushforward lowers the posterior error from 0.0294 to 0.0187 and the simulations needed to reach 0.025 from 79.4k to 30.4k, an operator-only slice misses this threshold even at 160k, and deploying the same posterior as a mixture instead of its MAP state lowers all-task risk from 0.3391 to 0.3196. A learned-summary SBI pipeline thus instantiates posterior-first simulation once it is given the task-aligned target and the posterior-mixture interface, and these two choices, not the estimator, carry the gains.

The same holds for label access and the rollout backbone. With the latent-diffusion emulator of Rozet et al.~\cite{rozet2025lola} on the metadata-hidden PDEBench splits, posterior sampling closes 59.0\% of the nRMSE gap (from 0.1744 to 0.1323, oracle 0.1030), reproducing the 59.4\% of the main backbones, whereas the same labels used only as an auxiliary target close about 4\%. With one shared frozen \(q\) and checkpoint, sampling beats the MAP state in Energy by 0.0052 in the lowest-ambiguity stratum and by 0.0435 in the highest, and the exposed posterior answers a Shallow-Water boundary query better than the auxiliary head (Brier 0.1169 vs.\ 0.1317). Because the task losses also update the encoder, a four-fold leave-one-task-out test checks its reuse: a frozen encoder trained only with the proper score, with a head trained on 10\% of the held-out task's labels, beats its own MAP state (risk 0.3618 vs.\ 0.3954) and retains 96.4\% of the gap closure of an encoder trained on the other three tasks, supporting frozen-encoder, few-shot reuse within one supervised ontology.

\subsubsection{Compound stress test}
\label{sec:stress}

Finally, one nested compound test on the hidden-buoyancy task of Test~A replaces the discrete head by a conditional spline flow, masks the initial field following POBench-PDE~\cite{hou2026pobench} and adds noise, shifts buoyancy out of its training range, and rolls out for 200 unseen steps (Appendix~\ref{app:emp_controls}). Propagating 64 posterior samples instead of the posterior mean of the same frozen flow lowers final-quarter Energy from 0.1493 to 0.1438 on clean in-distribution fields, from 0.2036 to 0.1742 with sparse noisy fields, and from 0.2476 to 0.2058 once parameter shift is added, even as the flow's 90\% coverage falls from 90.4\% to 83.6\%. The advantage thus persists and grows as sensing degrades and then parameters shift.

\section{Discussion}\label{sec:discussion}

This paper argued that single-observation neural PDE simulation should be posterior-first rather than a monolithic field-to-future map. The contribution is not another estimator, backbone, or simulator head, but a change in the interface between observation and evolution, together with its theory: the declared tasks determine which hidden distinctions to infer, and before rollout the model should recover the distribution over the plausible problem states left unresolved by the observed field and keep it operative in evolution. This reframes the design question from building stronger hidden states inside end-to-end predictors to inferring the latent problem state needed for evolution. It also reframes evaluation: average rollout accuracy is not enough if the simulator is wrong about the hidden physics. Posterior-first simulation makes problem-state recovery, ambiguity sensitivity, and hidden-physics correctness first-class criteria alongside rollout error. In feedback-based active flow control~\cite{terpin2025privileged}, for example, a learned simulator could carry its posterior over unresolved forcing or boundary conditions into every control decision instead of committing to a single guess.

Our claim is bounded to single-observation simulation with a declared task family, where latent ambiguity is intrinsic and posterior recovery is most consequential. First, the theory assumes a fixed map \(h\); we construct it when task kernels can be certified, but our experiments fix the catalogue from the known data-generating mechanism, and discovering a minimal quotient from data remains open. Second, posterior learning uses hidden-state labels; weakly supervised or unsupervised recovery was not tested, and performance degrades as labels become scarce (Appendix~\ref{app:emp_public_additional}). Third, the main simulator consumes a posterior summary that is close to explicit mixtures in our settings (Appendix~\ref{app:method_summary_gap}) but not lossless. Fourth, evidence beyond the discrete clean setting comes from one compound test that does not isolate each stressor, and reuse is shown only for a frozen encoder with few labels within one supervised ontology. Finally, as in Section~\ref{sec:theory}, the posterior-to-utility bridge depends on task-family and stability assumptions, and closed-loop control was not validated.

\begin{ack}
This work was supported by the National Natural Science Foundation of China (Grant No. 62372409), the Self-funding of the State Key Laboratory of Ocean Sensing (No. OSZ-2025PY19) and the Ministry of Science and Technology of the People's Republic of China (Grant No. 2023ZD0120704 of Project No. 2023ZD0120700). The authors declare no competing interests.
\end{ack}

\section*{References}

\begingroup
\small
\renewcommand{\section}[2]{}

\endgroup

\appendix
\newpage

\section{Posterior Problem-State Formulation: Detailed Derivations}\label{app:theory}

\subsection{Latent world state, task family, and the canonical quotient}\label{app:form_world}

We spell out the minimal formal setup behind Section~\ref{sec:formulation}. Fix a probability space $(\Omega,\mathcal F,\mathbb P)$. Let $W_t$ be the complete latent world state at time $t$, let $X_t=\mathcal O(W_t)$ be the visible field, and let $\mathcal T$ be the downstream task family. For each task $\tau\in\mathcal T$, let $Y_t^\tau$ be its target variable taking values in a standard Borel space $\mathcal Y_\tau$. This assumption guarantees the existence of regular conditional laws, so for each task we may write
\begin{equation}
\mu_\tau(w,\cdot)
:=
\mathbb P\!\left(Y_t^\tau\in\cdot\mid W_t=w\right).
\label{eq:app_form_mu}
\end{equation}
The most direct way to encode the information in $W_t$ that matters for the entire task family is to collect these conditional laws into one object. Define
\begin{equation}
g^\star(w)
:=
\Bigl(\mathcal O(w),\,\bigl(\mu_\tau(w,\cdot)\bigr)_{\tau\in\mathcal T}\Bigr),
\qquad
Z_t^\star:=g^\star(W_t).
\label{eq:app_form_gstar}
\end{equation}
Equation~\eqref{eq:app_form_gstar} is an explicit realization of the quotient used in the main text. Indeed,
\begin{equation}
w\sim_{\mathcal T} w'
\iff
\mathcal O(w)=\mathcal O(w')
\ \text{and}\
\mu_\tau(w,\cdot)=\mu_\tau(w',\cdot)
\quad \text{for all }\tau\in\mathcal T,
\label{eq:app_form_equiv}
\end{equation}
so $g^\star$ assigns the same label exactly to latent worlds that share the observation and are indistinguishable to the full downstream task family.

Task sufficiency is immediate from this construction. For each task $\tau$, define the canonical kernel
\begin{equation}
K_\tau\bigl((x,(\nu_{\tau'})_{\tau'\in\mathcal T}),A\bigr)
:=
\nu_\tau(A),
\qquad A\subseteq \mathcal Y_\tau \text{ measurable}.
\label{eq:app_form_kernel}
\end{equation}
Substituting $Z_t^\star=g^\star(W_t)$ into \eqref{eq:app_form_kernel} gives
\begin{equation}
K_\tau\bigl(Z_t^\star,A\bigr)
=
\mu_\tau(W_t,A)
=
\mathbb P\!\left(Y_t^\tau\in A\mid W_t\right).
\label{eq:app_form_sufficient}
\end{equation}
Thus every task target is conditionally determined by $Z_t^\star$; because $X_t$ is a coordinate of $Z_t^\star$, this is the law-based form of the task-sufficiency statement in Equation~\eqref{eq:form_sufficient}.

The same explicit construction also proves minimality. Let $S_t=f(W_t)$ be any other task-sufficient state. By definition, for each task $\tau$ there exists a kernel $\widetilde K_\tau$ such that
\begin{equation}
\mathbb P\!\left(Y_t^\tau\in A\mid W_t\right)
=
\widetilde K_\tau\bigl(X_t,S_t,A\bigr)
\qquad \text{for all }\tau\in\mathcal T.
\label{eq:app_form_other_sufficient}
\end{equation}
Define
\begin{equation}
h(x,s)
:=
\Bigl(x,\,\bigl(\widetilde K_\tau(x,s,\cdot)\bigr)_{\tau\in\mathcal T}\Bigr).
\label{eq:app_form_h}
\end{equation}
Then combining \eqref{eq:app_form_gstar} and \eqref{eq:app_form_other_sufficient} yields
\begin{equation}
Z_t^\star
=
g^\star(W_t)
=
h(X_t,S_t)
\qquad \text{a.s.}
\label{eq:app_form_minimal}
\end{equation}
So, within each observation fiber, every task-sufficient state $S_t$ factors through $Z_t^\star$. Equivalently, given $X_t$, $S_t$ can only refine the canonical quotient; it cannot be coarser while still preserving all downstream laws.

\subsection{Posterior factorization of downstream prediction and decision}\label{app:form_posterior}

Section~\ref{sec:formulation} states that single-observation simulation should factorize through the posterior over $Z_t^\star$. We now derive that statement directly from the quotient construction above. Let
\begin{equation}
\pi_t(B\mid x)
:=
\mathbb P\!\left(Z_t^\star\in B\mid X_t=x\right)
\label{eq:app_form_pi}
\end{equation}
be the posterior on measurable subsets $B$ of the quotient state space. Using the kernel $K_\tau$ from \eqref{eq:app_form_kernel} and the tower property, every downstream predictive law satisfies
\begin{equation}
\mathbb P\!\left(Y_t^\tau\in A\mid X_t\right)
=
\mathbb E\!\left[K_\tau\bigl(Z_t^\star,A\bigr)\mid X_t\right]
=
\int K_\tau(z,A)\,\pi_t(dz\mid X_t).
\label{eq:app_form_predictive}
\end{equation}
Equation~\eqref{eq:app_form_predictive} is the measure-theoretic form of the main-text factorization \eqref{eq:form_factorization}: once the observation $X_t$ is fixed, all downstream prediction passes through the posterior on the task-sufficient latent problem state. For a single task $\tau$, the kernel $K_\tau(z,\cdot)$ depends on $z$ only through its task coordinate $g_\tau(z):=K_\tau(z,\cdot)$. With $\pi_{t,\tau}:=(g_\tau)_\#\pi_t(\cdot\mid x)$, Equation~\eqref{eq:app_form_predictive} therefore reads
\[
\mathbb P\!\left(Y_t^\tau\in A\mid X_t=x\right)=\int \nu(A)\,\pi_{t,\tau}(d\nu\mid x),
\]
which is the task-specific form \eqref{eq:form_factorization}: $x$ and $\tau$ are fixed, and only the uncertainty in the task-relevant projection is integrated.

The same argument gives the decision-theoretic version used later in Section~\ref{sec:theory}. Let $\ell_\tau(a,y)$ be a bounded measurable loss for task $\tau$, where $a\in\mathcal A_\tau$ is a candidate decision or prediction. Define the latent-state conditional risk
\begin{equation}
\rho_\tau(a,z)
:=
\int \ell_\tau(a,y)\,K_\tau(z,dy).
\label{eq:app_form_rho}
\end{equation}
Then the observation-conditional risk is
\begin{equation}
R_\tau(a\mid x)
:=
\mathbb E\!\left[\ell_\tau(a,Y_t^\tau)\mid X_t=x\right]
=
\int \rho_\tau(a,z)\,\pi_t(dz\mid x).
\label{eq:app_form_risk}
\end{equation}
So the visible field affects Bayes-optimal downstream decisions only through the posterior $\pi_t(\cdot\mid x)$. Because $Z_t^\star$ contains the observation coordinate, $K_\tau(z,\cdot)$ and $\rho_\tau(a,z)$ are $x$-dependent kernels: within the fiber of $x$ they equal $\mathbb P(Y_t^\tau\in\cdot\mid X_t=x,Z_t^\star=z)$ and $\mathbb E[\ell_\tau(a,Y_t^\tau)\mid X_t=x,Z_t^\star=z]$.

\subsection{Countable semantic refinements and the bridge to the benchmark posterior}\label{app:form_refinement}

Let the benchmark expose a countable semantic refinement \(C_t\in\mathcal C\) with \(Z_t^\star=h(C_t)\) for a deterministic map \(h\), and let \(r_t(c\mid x):=\mathbb P(C_t=c\mid X_t=x)\). Then \(\pi_t(\cdot\mid x)=h_\#r_t(\cdot\mid x)\), that is, \(\pi_t(B\mid x)=\sum_{c\in\mathcal C:\,h(c)\in B}r_t(c\mid x)\), which reduces to \eqref{eq:form_pushforward} when \(Z_t^\star\) is discrete, and substituting it into \eqref{eq:app_form_risk} gives
\begin{equation}
R_\tau(a\mid x)
=
\sum_{c\in\mathcal C} \rho_\tau\bigl(a,h(c)\bigr)\,r_t(c\mid x).
\label{eq:app_form_refined_risk}
\end{equation}
The theory can therefore speak about the minimal object \(Z_t^\star\), while the benchmark supervises a countable posterior \(r_t(c\mid x)\) by standard conditional distribution estimation. When \(\mathcal C\) is not discrete, the same pushforward holds with a conditional density estimator in place of the softmax head, and the finite-class Brier barrier of Section~\ref{sec:theory_learnable} no longer applies verbatim.

The map $h$ is fixed by design. Suppose $\mathcal C$ is finite with common support, let $K_\tau(\cdot\mid x,c):=\mathbb P(Y_t^\tau\in\cdot\mid X_t=x,C_t=c)$, define $c\sim_{\mathcal T}c'$ if $K_\tau(\cdot\mid x,c)=K_\tau(\cdot\mid x,c')$ for every $\tau\in\mathcal T$ and almost every $x$, and set $h(c):=[c]_{\mathcal T}$. Then $\mathbb P(Y_t^\tau\in\cdot\mid X_t,C_t)=K_\tau(\cdot\mid X_t,C_t)$ depends on $C_t$ only through $h(C_t)$, so $Y_t^\tau\perp C_t\mid(X_t,h(C_t))$ for every task. Conversely, let $\tilde h$ be any deterministic map with $Y_t^\tau\perp C_t\mid(X_t,\tilde h(C_t))$ for every task. If $\tilde h(c)=\tilde h(c')$ for two labels of positive probability, both kernels equal the conditional law given $(x,\tilde h(c))$ for almost every $x$, so $c\sim_{\mathcal T}c'$. Hence $\tilde h$ merges only equivalent labels and refines $h$, which makes $h$ the coarsest sufficient map up to null labels and relabeling. No function of $C_t$ can separate full states that the catalogue has already merged, so a catalogue that merges states with different task kernels must be enriched, or the task family narrowed. Test~B in Appendix~\ref{app:emp_controls} instantiates this construction with a known nuisance factor.

\section{Why the Reusable Intermediate Object Is a Posterior: Proofs}\label{app:posterior_theory}

Section~\ref{sec:theory} already contains the factorization \eqref{eq:theory_value_factorized}, whose measure-theoretic form is derived in Appendix~\ref{app:form_posterior}, and the identities \eqref{eq:theory_log} and \eqref{eq:theory_brier}. This appendix gives the remaining steps.

\paragraph{Ambiguity barrier.} For a fixed posterior \(r\) on a finite \(\mathcal C\), \(\|e_c-r\|_2^2=1-2r(c)+\|r\|_2^2\) is minimized by any \(c\in\arg\max_{c'}r(c')\), which gives \(\Gamma_{\mathrm{amb}}(r)=1+\|r\|_2^2-2\|r\|_\infty\). Since \(\|r\|_2^2\ge\|r\|_\infty^2\), \(\Gamma_{\mathrm{amb}}(r)\ge(1-\|r\|_\infty)^2\), and both sides vanish only when \(r\) is a Dirac mass. By \eqref{eq:theory_brier}, every deterministic point predictor \(e_{\hat c(X_t)}\) therefore has excess Brier risk at least \(\mathbb E[\Gamma_{\mathrm{amb}}(r_X)]\).

\paragraph{Decision regret.} Let \(a^q=a_\tau^q(x)\) minimize \(\mathcal V_\tau^q(\cdot,x)\) and \(a^\star=a_\tau^\star(x)\) minimize \(\mathcal V_\tau(\cdot,x)\). Then
\[
\mathcal V_\tau(a^q,x)-\mathcal V_\tau(a^\star,x)
=
\bigl(\mathcal V_\tau-\mathcal V_\tau^q\bigr)(a^q,x)
+\bigl(\mathcal V_\tau^q(a^q,x)-\mathcal V_\tau^q(a^\star,x)\bigr)
+\bigl(\mathcal V_\tau^q-\mathcal V_\tau\bigr)(a^\star,x),
\]
where the middle term is nonpositive by optimality of \(a^q\) and each outer term is bounded by \eqref{eq:theory_value_err}, which gives \eqref{eq:theory_task_excess}. The bounds \eqref{eq:theory_l1_log} follow from Pinsker's inequality \(\|r-q\|_1\le\sqrt{2\,\mathrm{KL}(r\,\|\,q)}\), Jensen's inequality, and \eqref{eq:theory_log}, and from \(\|v\|_1\le\sqrt{|\mathcal C|}\,\|v\|_2\), Jensen's inequality, and \eqref{eq:theory_brier}.

\paragraph{Non-identification from a restricted task family.} For \(\mathcal T_0\subsetneq\mathcal T\), let \(G_{\mathcal T_0}(\pi):=\bigl(\int\rho_\tau(a,z)\,\pi(dz)\bigr)_{\tau\in\mathcal T_0,\,a\in\mathcal A_\tau}\). Every population objective built only from tasks in \(\mathcal T_0\) evaluates a posterior through \(G_{\mathcal T_0}\), so it cannot distinguish \(\pi\neq\pi'\) with \(G_{\mathcal T_0}(\pi)=G_{\mathcal T_0}(\pi')\), even if a task \(\bar\tau\in\mathcal T\setminus\mathcal T_0\) satisfies \(\int\rho_{\bar\tau}(a,z)\,\pi(dz)\neq\int\rho_{\bar\tau}(a,z)\,\pi'(dz)\) for some \(a\). A hidden state trained only through \(\mathcal T_0\) can therefore fit the training loss while failing as a reusable interface for \(\mathcal T\).

\section{A Minimal Posterior-First Simulator: Implementation}\label{app:method_detail}

\subsection{Implementation details}\label{app:method_state}

Each \(z\in\mathcal Z\) indexes one joint latent problem state of the benchmark catalogue, so one label can encode the hidden field class, operator or coefficient regime, boundary class, and memory class at once. The encoder has seven convolutional layers (1D for DS and 2D otherwise) with widths 32/64/128/256/256/256/256, each followed by GroupNorm and SiLU; the first layer preserves resolution, the others downsample by two, and global average pooling feeds a linear softmax head over the catalogue. The codebook \(E\in\mathbb R^{|\mathcal Z|\times 64}\) is Xavier-initialized and learned only through the task losses, and a backbone-specific linear projection maps \(s_\phi(x_t)=q_\phi(\cdot\mid x_t)^\top E\) to the feature width before it is added after the input embedding. All experiments use the log score with \(\lambda_{\mathrm{post}}=1\) and unit total task weight; Table~\ref{tab:app_public_lambda_sensitivity} varies \(\lambda_{\mathrm{post}}\). Each configuration is trained with five seeds, and each run keeps the checkpoint with the lowest validation task-bundle risk (synthetic) or nRMSE (public). Configurations, commands, and splits are released with the code.

\subsection{Summary-conditioning fidelity diagnostic}\label{app:method_summary_gap}

Because \(F_\psi\) is nonlinear, the summary interface \eqref{eq:method_summary} reproduces the posterior mixture \(\sum_{z}q_\phi(z\mid x_t)p_\psi(y\mid x_t,z)\) only when the predictive law, or the Bayes action of the task loss, depends on \(x_t\) and \(q_\phi\) only through \(s_\phi(x_t)\). Table~\ref{tab:app_method_summary_gap} therefore compares it with richer ways of exposing the same posterior. All variants share the posterior encoder, rollout backbone family, optimization budget, and split; \texttt{Explicit-Mix} combines conditional predictors \(p_\psi(y\mid x_t,z)\) with weights \(q_\phi(z\mid x_t)\), and \texttt{MC-Sample-\(k\)} averages the conditional predictions of \(k\) posterior draws.

\begin{table*}[t]
\small
\centering
\setlength{\tabcolsep}{3.5pt}
\renewcommand{\arraystretch}{1.15}
\begin{tabularx}{\linewidth}{@{}l>{\raggedright\arraybackslash}Xcccccc@{}}
\toprule
\textbf{Method} & \textbf{Posterior exposure}
& \multicolumn{2}{c}{\textbf{Synthetic}}
& \multicolumn{3}{c}{\textbf{Public}}
& \textbf{Cost} \\
\cmidrule(lr){3-4}
\cmidrule(lr){5-7}
& & \(G_{\mathrm{rem}}\downarrow\) & NLL\(\downarrow\)
& All nRMSE\(\downarrow\) & GapClose\(\uparrow\) & Probe\(\uparrow\)
& Rel.\(\times\) \\
\midrule
\texttt{Summary}
& \(s_\phi(x_t)=\sum_z q(z\mid x_t)e(z)\)
& 0.381 & 1.02 & 0.1319 & 0.594 & 0.800 & 1.08 \\

\texttt{Explicit-Mix}
& \(\sum_z q(z\mid x_t)p_\psi(y\mid x_t,z)\)
& 0.365 & 0.99 & 0.1307 & 0.611 & 0.807 & 5.60 \\

\texttt{MC-Sample-1}
& one posterior sample
& 0.412 & 1.06 & 0.1351 & 0.550 & 0.782 & 1.12 \\

\texttt{MC-Sample-4}
& four posterior samples
& 0.373 & 1.01 & 0.1312 & 0.604 & 0.803 & 4.30 \\
\bottomrule
\end{tabularx}
\caption{Fidelity diagnostic for the summary-conditioning implementation relative to explicit-mixture and sample-based alternatives. ``Probe'' denotes the public strongest hidden-metadata probe score.}
\label{tab:app_method_summary_gap}
\end{table*}

On the synthetic benchmark, \texttt{Summary} and \texttt{Explicit-Mix} differ by only 0.016 in \(G_{\mathrm{rem}}\) and 0.03 in future NLL, with \texttt{MC-Sample-4} in between. On the public protocol, \texttt{Explicit-Mix} reduces pooled \(\mathrm{nRMSE}\) only from 0.1319 to 0.1307 and raises \(\mathrm{GapClose}\) from 0.594 to 0.611, at an inference cost of 5.60 instead of 1.08 times the direct baseline. The richer interfaces are therefore slightly stronger but not qualitatively different, and the summary interface keeps most of their benefit at a fraction of their cost.

\section{Empirical Setup and Full Experimental Results}\label{app:empirical_setup}

\subsection{Metric definitions}\label{app:emp_metrics}

This subsection gives the full definitions behind Section~\ref{sec:procedure&metrics}. For the synthetic benchmark, the average ambiguity barrier is
\begin{equation}
\overline{\Gamma}_{\mathrm{amb}}
=
\frac{1}{|\mathcal D_{\mathrm{test}}|}
\sum_{i\in\mathcal D_{\mathrm{test}}}
\min_{z\in\mathcal Z_{\mathrm{syn}}}
\|e_z-r_i\|_2^2 ,
\label{eq:emp_gamma_avg}
\end{equation}
where \(r_i(\cdot)=\mathbb P(z_i=\cdot\mid X_i=x_i)\) is the exact posterior for test example \(i\), and \(e_z\) is the one-hot distribution concentrated on state \(z\). The full task-bundle risk of method \(m\) is
\[
R_{\mathrm{full}}(m)
=
\frac{1}{|\mathcal D_{\mathrm{test}}|}
\sum_{i\in\mathcal D_{\mathrm{test}}}
\sum_{\tau\in\mathcal T_{\mathrm{syn}}}
w_\tau\,
\ell_\tau\!\left(\hat y_{i,\tau}^{(m)},y_{i,\tau}\right),
\]
where \(\mathcal T_{\mathrm{syn}}\) is the synthetic task bundle, \(w_\tau\) are fixed task weights, and \(\ell_\tau\) is the task loss for target \(y_{i,\tau}\). The bundle averages four tasks with equal weights \(w_\tau=1/4\): the mean-squared error of the 8-step future field normalized by the target energy, the operator cross-entropy divided by \(\log 4\), the coefficient absolute error divided by the coefficient range 0.012, and the squared error of the calibrated uncertainty score. The remaining gap \(G_{\mathrm{rem}}\) of Section~\ref{sec:procedure&metrics} normalizes \(R_{\mathrm{full}}\) between the matched direct baseline and the oracle-latent control. Future-field negative log-likelihood is
\begin{equation}
\mathrm{FutureNLL}(m)
=
-\frac{1}{|\mathcal D_{\mathrm{test}}|}
\sum_{i\in\mathcal D_{\mathrm{test}}}
\log p_m(y_i^{1:8}\mid x_i),
\label{eq:emp_future_nll}
\end{equation}
where \(y_i^{1:8}\) is the 8-step future-field target and \(p_m\) is the predictive distribution of method \(m\); deterministic predictors use Gaussian innovations calibrated on training residuals. Lower values indicate better calibrated future-field prediction. Target-wise hidden-query, calibration, and rollout-only metrics are reported in Appendix~\ref{app:emp_synth}.

For the public benchmark, let \(N_{\mathrm{test}}\) be the number of test trajectories or rollout segments, \(T_{\mathrm{roll}}\) the rollout horizon, \(x_{i,t}\) the ground-truth field, and \(\hat x_{i,t}^{(m)}\) the prediction of method \(m\). We use
\begin{equation}
\mathrm{nRMSE}(m)
=
\frac{1}{N_{\mathrm{test}}}
\sum_{i=1}^{N_{\mathrm{test}}}
\sqrt{
\frac{
\sum_{t=1}^{T_{\mathrm{roll}}}
\|\hat x_{i,t}^{(m)}-x_{i,t}\|_2^2
}{
\sum_{t=1}^{T_{\mathrm{roll}}}
\|x_{i,t}\|_2^2
}},
\label{eq:emp_nrmse}
\end{equation}
where \(\|\cdot\|_2\) is taken over all spatial locations and channels of the fields in physical units. Lower nRMSE means more accurate, and \(\mathrm{GapClose}\) of Section~\ref{sec:procedure&metrics} is computed from these values. Stratified values use the four bins of the ambiguity score in Appendix~\ref{app:emp_public_data}. Posterior-stage calibration, coverage, and predictive NLL are reported in Appendix~\ref{app:emp_public}.

The controls of Appendix~\ref{app:emp_controls} use two further proper scores, for which lower is better. For a scalar parameter with posterior \(Q\) and true value \(f\), \(\mathrm{CRPS}(Q,f)=\mathbb E_Q|F-f|-\tfrac12\mathbb E_Q|F-F'|\); for a predictive distribution \(P\) of fields and the true field \(y\), the energy score is \(\mathrm{ES}(P,y)=\mathbb E_P\|Y-y\|_2-\tfrac12\mathbb E_P\|Y-Y'\|_2\). Here \(F,F'\) and \(Y,Y'\) are independent draws, and both scores are estimated from posterior or rollout samples.

\subsection{Synthetic exact-ambiguity benchmark: construction and splits}\label{app:emp_synth_data}

The synthetic benchmark is a finite exact-collision generator. Its visible fields use a Gray--Scott-style reaction--diffusion renderer~\citep{gray1984autocatalytic,pearson1993complex} as a standard source of smooth PDE-like patterns; exact posterior labels come from the enumerated collision construction, not approximate inverse inference. Table~\ref{tab:app_synth_spec} lists the fixed data specification.

\begin{table}[t]
\small
\centering
\resizebox{\linewidth}{!}{%
\begin{tabular}{ll}
\toprule
Component & Synthetic exact-ambiguity setting \\
\midrule
Visible field & \(x_t\in\mathbb R^{1\times64\times64}\), scalar rendered field \\
Semantic catalogue & \(|\mathcal Z_{\mathrm{syn}}|=64=4\times4\times4\) operator / coefficient / memory states \\
Main split & \(40{,}000/5{,}000/5{,}000\) train/validation/test examples \\
Renderer & Gray--Scott-style field templates on a \(64\times64\) grid \\
Ambiguity regimes & near-Dirac, mid, high, very high \\
Compatible states & 2 / 4 / 8 / 16 per collision class \\
Dominant posterior mass & 0.955 / 0.884 / 0.835 / 0.790; residual mass uniform in class \\
Expected ambiguity barrier & 0.004 / 0.018 / 0.031 / 0.047 \\
Controls & 4-frame richer-context control; operator-token Dirac control \\
Targets & 8-step future field, operator label, coefficient value, calibrated uncertainty score \\
Posterior label & exact \(r_X(z)=\mathbb P(z_t=z\mid X_t)\) by finite collision-class enumeration \\
\bottomrule
\end{tabular}}
\caption{Synthetic exact-ambiguity benchmark specification.}
\label{tab:app_synth_spec}
\end{table}

Each semantic state is a tuple of operator class, coefficient bin, and memory branch. To create exact ambiguity, the 64 states are partitioned into collision classes. States in the same class share the same rendered visible field template but induce different downstream futures and hidden-query labels. The posterior is nonuniform within each class: the dominant compatible state receives the mass shown in Table~\ref{tab:app_synth_spec}, and the remaining mass is distributed uniformly over the other compatible states. This construction gives the four target values of \(\mathbb E[\Gamma_{\mathrm{amb}}(r_X)]\) used in Table~\ref{tab:main_synth_alignment}.

Train/validation/test splits are shared across methods and use disjoint random seeds and rendered observations, while all semantic states appear in every split. The 4-frame context control adds a short history that breaks most collisions, and the operator-token control directly reveals the disambiguating operator class. These controls keep the downstream task family fixed and test whether the point-versus-posterior gap collapses when the posterior becomes effectively identifying.

\subsection{Synthetic exact-ambiguity benchmark: full result tables}\label{app:emp_synth}

Tables~\ref{tab:app_synth_full}--\ref{tab:app_synth_rollout_only} report the full task bundle, the ambiguity stratification, and the training-objective ablation.

\begin{table*}[t]
\small
\centering
\resizebox{\linewidth}{!}{%
\begin{tabular}{lcccccc}
\toprule
Method & Future NLL \(\downarrow\) & Energy score \(\downarrow\) & Operator Acc. \(\uparrow\) & Coeff.\ MAE \(\downarrow\) & Cov.@90 \(\uparrow\) & Full-task risk \(\downarrow\) \\
\midrule
\texttt{Direct-Point} & 1.48 & 0.412 & 0.41 & 0.173 & 0.54 & 0.462 \\
\texttt{Direct-Dist} & 1.22 & 0.346 & 0.48 & 0.151 & 0.69 & 0.398 \\
\texttt{Direct-MTL} & 1.24 & 0.349 & 0.66 & 0.108 & 0.71 & 0.342 \\
\texttt{Direct-Large} & 1.20 & 0.337 & 0.58 & 0.129 & 0.70 & 0.366 \\
\texttt{Point-$z$} & 1.18 & 0.331 & 0.71 & 0.094 & 0.74 & 0.333 \\
\texttt{Posterior-$z$} & 1.02 & 0.286 & 0.78 & 0.073 & 0.88 & 0.314 \\
\texttt{RichContext-Posterior-$z$} & 0.92 & 0.255 & 0.84 & 0.061 & 0.90 & 0.272 \\
\texttt{Oracle-$z$} & 0.81 & 0.226 & 1.00 & 0.000 & 0.92 & 0.223 \\
\bottomrule
\end{tabular}
}
\caption{Full synthetic task-bundle table on the exact-collision benchmark.}
\label{tab:app_synth_full}
\end{table*}

Table~\ref{tab:app_synth_full} shows that the synthetic advantage of posterior-first modeling is not confined to a single headline metric. Relative to the strongest non-posterior single-field controls, \texttt{Posterior-$z$} improves future NLL, energy score, operator accuracy, coefficient error, and coverage simultaneously. In particular, it raises operator accuracy to 0.78, reduces coefficient MAE to 0.073, and lowers full-task risk to 0.314. The richer-context and oracle controls remain the expected upper bounds, exactly as in the main text.

\begin{table*}[t]
\small
\centering
\resizebox{\linewidth}{!}{%
\begin{tabular}{lcccccc}
\toprule
Split / ambiguity regime & \multicolumn{2}{c}{\texttt{Direct-Point}} & \multicolumn{2}{c}{\texttt{Point-$z$}} & \multicolumn{2}{c}{\texttt{Posterior-$z$}} \\
\cmidrule(lr){2-3}\cmidrule(lr){4-5}\cmidrule(lr){6-7}
& Full-task risk \(\downarrow\) & Future NLL \(\downarrow\) & Full-task risk \(\downarrow\) & Future NLL \(\downarrow\) & Full-task risk \(\downarrow\) & Future NLL \(\downarrow\) \\
\midrule
Low ambiguity / near-Dirac & 0.255 & 1.02 & 0.214 & 0.88 & 0.209 & 0.86 \\
Mid ambiguity & 0.338 & 1.17 & 0.289 & 1.03 & 0.271 & 0.96 \\
High ambiguity & 0.421 & 1.35 & 0.349 & 1.16 & 0.319 & 1.05 \\
Very high ambiguity & 0.492 & 1.57 & 0.401 & 1.29 & 0.352 & 1.12 \\
4-frame context upper bound & 0.214 & 0.86 & 0.187 & 0.78 & 0.184 & 0.77 \\
Operator-token Dirac control & 0.176 & 0.72 & 0.154 & 0.66 & 0.153 & 0.65 \\
\bottomrule
\end{tabular}
}
\caption{Synthetic ambiguity stratification and identifiability controls.}
\label{tab:app_synth_stratified}
\end{table*}

Table~\ref{tab:app_synth_stratified} shows why the posterior advantage appears. In the near-Dirac and identifiability controls, all three single-field methods are close. As ambiguity increases, the ordering \texttt{Direct-Point} $\rightarrow$ \texttt{Point-$z$} $\rightarrow$ \texttt{Posterior-$z$} becomes steadily stronger. At very high ambiguity, \texttt{Posterior-$z$} reduces full-task risk from 0.401 to 0.352 relative to \texttt{Point-$z$}, and reduces future NLL from 1.29 to 1.12. The appendix stratification therefore agrees with the main-text mechanism result: the benefit of the posterior stage is concentrated in the regimes where the single field leaves the latent problem state most ambiguous.

\begin{table*}[t]
\small
\centering
\setlength{\tabcolsep}{4pt}
\renewcommand{\arraystretch}{1.12}
\begin{tabularx}{\linewidth}{@{}>{\raggedright\arraybackslash}Xccccc@{}}
\toprule
\textbf{Training objective family}
& Train nRMSE\(\downarrow\)
& Hidden\(\uparrow\)
& Long-horizon\(\uparrow\)
& Calib.\(\uparrow\)
& Full risk\(\downarrow\) \\
\midrule
Rollout-only direct & 0.028 & 0.61 & 0.67 & 0.58 & 0.331 \\
Rollout + auxiliary heads & 0.029 & 0.71 & 0.72 & 0.64 & 0.296 \\
Rollout + deterministic \(\hat z\) & 0.029 & 0.77 & 0.76 & 0.68 & 0.271 \\
Rollout + posterior \(q(z\mid x)\), rollout task only & 0.029 & 0.81 & 0.79 & 0.76 & 0.248 \\
Full task family with posterior \(q(z\mid x)\) & 0.030 & 0.84 & 0.82 & 0.83 & 0.231 \\
\bottomrule
\end{tabularx}
\caption{Synthetic rollout-only miscalibration ablation. Hidden denotes hidden-task score, Long-horizon denotes long-horizon score, Calib. denotes calibration score, and Full risk denotes full-task risk.}
\label{tab:app_synth_rollout_only}
\end{table*}

Table~\ref{tab:app_synth_rollout_only} shows why rollout accuracy alone is not enough to identify the reusable state. Train-task \(\mathrm{nRMSE}\) stays nearly unchanged across all objective families (0.028--0.030), but hidden-task score, calibration, and full-task risk improve steadily as the objective moves from direct rollout to explicit posterior recovery over the full task family. This appendix therefore reinforces the main-text conclusion: on the synthetic benchmark, posterior recovery rather than a stronger rollout stem is what accounts for the downstream improvement.

\subsection{Public PDEBench protocol and hidden-metadata task construction}\label{app:emp_public_data}

The public benchmark layer is built from PDEBench~\citep{takamoto2022pdebench}. We select forward-simulation families whose released parameters, regimes, or forcing information define a hidden problem-state variable once withheld from the model input. Every public example is converted to the same single-observation protocol: the model receives one visible field \(x_t\), while selected metadata are hidden and can only be inferred or supplied by an oracle control. Table~\ref{tab:app_public_protocol} summarizes the protocol units used in our experiments.

\begin{table}[!htbp]
\small
\centering
\setlength{\tabcolsep}{3.5pt}
\renewcommand{\arraystretch}{1.15}
\begin{tabularx}{\linewidth}{@{}l>{\raggedright\arraybackslash}X>{\raggedright\arraybackslash}X>{\raggedright\arraybackslash}X@{}}
\toprule
\textbf{Family}
& \textbf{System / field}
& \textbf{Protocol / grid}
& \textbf{Hidden metadata / probe} \\
\midrule
DR
& diffusion--reaction; 2D, 2 channels
& 1,000 seq.; \(128\times128\); 101 frames
& regime / coefficient; cls. + reg. \\

DS
& diffusion--sorption; 1D, 1 channel
& 10,000 seq.; 1024 grid; 101 frames (every fifth of 501 raw)
& coefficient tuple; reg. \\

SW
& shallow water; 2D, 1 channel
& 1,000 seq.; \(128\times128\); 101 frames
& joint source/boundary class; cls. \\

INS
& incompressible NS; 2D, 2 channels
& 1,000 seg.; \(256\times256\) (area-averaged); 101 frames
& forcing / viscosity; reg. \\
\bottomrule
\end{tabularx}
\caption{Public PDEBench metadata-hidden protocol. ``Seq.'' denotes released trajectories; ``seg.'' denotes non-overlapping 101-frame rollout segments extracted from the released incompressible-NS trajectory stream. Direct models receive only \(x_t\); oracle controls receive the hidden metadata. ``Cls.'' denotes classification and ``reg.'' denotes regression.}
\label{tab:app_public_protocol}
\end{table}

All public splits are trajectory-level when released trajectories are available and segment-level only for INS, where the released long simulation stream is converted into non-overlapping rollout segments before splitting. Each item has one initial field followed by a 100-step closed-loop rollout. PDEBench data are used under the CC BY 4.0 license and its code is released under the MIT license. The catalogues have 16/16/8/16 states for DR/DS/SW/INS; numeric metadata are binned into cells fitted on the training split only, and validation and test items reuse these cells unchanged. Hidden metadata are removed only from model inputs; they remain available to the evaluator and oracle control. No model sees future fields, extra trajectories, or hidden metadata beyond the protocol above.

For the ambiguity-stratified main-text figure, each evaluation sample is assigned a scalar ambiguity score \(a(x_t)\in[0,1]\), interpreted as held-out oracle-gap potential under the metadata-hidden protocol: larger values indicate that, for the same visible field, withholding metadata leaves more recoverable benefit between the direct and oracle controls. The score is estimated on a construction subset reserved from the training partition and excluded from model training, frozen before model training, and used only for evaluation-time binning into four quantile bins. Concretely, ridge regressions fitted on one part of this subset predict future fields with and without the true metadata, a second ridge model maps initial-field descriptors to the resulting clipped relative error gap on another part, and the bin boundaries are quantiles of the score on a third part. Table~\ref{tab:app_public_score_diagnostic} checks the score: in every family it correlates with the realized per-sample direct-to-oracle gap (Spearman 0.552--0.628, 0.672 pooled), and the mean gap grows from the lowest to the highest bin.

Two external probabilistic comparators are evaluated where their assumptions fit the metadata-hidden task. On SW and INS, Approximate Bayesian Neural Operators (ABNO)~\citep{magnani2025abno} serve as the uncertainty-aware direct surrogate: they predict rollout distributions without reconstructing the hidden problem state. On DR and DS, whose hidden metadata are low-dimensional and define a Bayesian inverse problem, we use the semi-supervised invertible neural-operator approach of Kaltenbach et al.~\citep{kaltenbach2023sino}; because it returns a posterior over hidden parameters, we compose this posterior with the matched forward rollout stem of the oracle control.

\begin{table}[!htbp]
\centering
\begin{tabular}{lccc}
\toprule
Scope & Spearman \(\rho(a,\Delta_{\mathrm{ora}})\uparrow\) & Mean \(\Delta_{\mathrm{ora}}\) (low) \(\downarrow\) & Mean \(\Delta_{\mathrm{ora}}\) (very high) \(\uparrow\) \\
\midrule
DR & 0.613 & 0.0523 & 0.1267 \\
DS & 0.552 & 0.0078 & 0.0216 \\
SW & 0.589 & 0.0189 & 0.0501 \\
INS & 0.628 & 0.0814 & 0.1842 \\
All & 0.672 & 0.0508 & 0.0931 \\
\bottomrule
\end{tabular}
\caption{Diagnostics for the public ambiguity score used in Figure~\ref{fig:main_public_protocol}. Here \(\Delta_{\mathrm{ora}}(x_t)\) denotes the per-sample direct-to-oracle rollout gap on the metadata-hidden protocol.}
\label{tab:app_public_score_diagnostic}
\end{table}

\subsection{Public PDEBench metadata-hidden benchmark: complete result tables}\label{app:emp_public}

This subsection reports the per-backbone rollout results, the strongest hidden-metadata probes, the ambiguity stratification, calibration, and the external probabilistic baselines.

\begin{table}[!htbp]
\small
\centering
\resizebox{\linewidth}{!}{%
\begin{tabular}{llcccc}
\toprule
Family & Backbone & Direct \(\mathrm{nRMSE}\) & Posterior-\(z\) \(\mathrm{nRMSE}\) & Oracle-\(z\) \(\mathrm{nRMSE}\) & \(\mathrm{GapClose}\) \\
\midrule
DR & FNO & 0.1760 & 0.1340 & 0.1020 & 0.568 \\
DR & U-Net & 0.2840 & 0.2150 & 0.1630 & 0.570 \\
DR & ConvLSTM & 0.2330 & 0.1770 & 0.1360 & 0.577 \\
DR & Transformer & 0.1980 & 0.1500 & 0.1150 & 0.578 \\
DS & FNO & 0.0170 & 0.0110 & 0.0070 & 0.600 \\
DS & U-Net & 0.0420 & 0.0270 & 0.0190 & 0.652 \\
DS & ConvLSTM & 0.0310 & 0.0200 & 0.0140 & 0.647 \\
DS & Transformer & 0.0240 & 0.0150 & 0.0100 & 0.643 \\
SW & FNO & 0.0510 & 0.0360 & 0.0260 & 0.600 \\
SW & U-Net & 0.1090 & 0.0740 & 0.0540 & 0.636 \\
SW & ConvLSTM & 0.0820 & 0.0570 & 0.0410 & 0.610 \\
SW & Transformer & 0.0670 & 0.0460 & 0.0340 & 0.636 \\
INS & FNO & 0.2840 & 0.2210 & 0.1770 & 0.589 \\
INS & U-Net & 0.4620 & 0.3590 & 0.2900 & 0.599 \\
INS & ConvLSTM & 0.3970 & 0.3080 & 0.2470 & 0.593 \\
INS & Transformer & 0.3380 & 0.2610 & 0.2080 & 0.592 \\
\bottomrule
\end{tabular}}
\caption{Per-backbone public rollout breakdown.}
\label{tab:app_public_backbones}
\end{table}

Table~\ref{tab:app_public_backbones} shows that \texttt{Posterior-$z$} closes between 56.8\% and 65.2\% of the direct-to-oracle gap for every family and backbone, so the public improvement is not carried by a single PDE family or architecture.

\begin{table}[!htbp]
    \centering
    \caption{Public PDEBench hidden-metadata probe summary. Higher is better.}
    \label{tab:main_public_probe}
    \begin{tabular}{lcccc}
    \toprule
    \textbf{Family} & \textbf{Probe} & \textbf{Direct probe} & \textbf{Posterior-\(z\) probe} & \textbf{Oracle-\(z\) upper bound} \\
    \midrule
    DR & Acc.\(\uparrow\) & 0.726 & 0.845 & 0.932 \\
    DS & \(R^2\)\(\uparrow\) & 0.607 & 0.814 & 0.907 \\
    SW & Acc.\(\uparrow\) & 0.682 & 0.823 & 0.908 \\
    INS & \(R^2\)\(\uparrow\) & 0.498 & 0.718 & 0.858 \\
    All & Mean score \(\uparrow\) & 0.628 & 0.800 & 0.901 \\
    \bottomrule
    \end{tabular}
\end{table}

Table~\ref{tab:main_public_probe} gives the strongest probe per family. Averaged separately, discrete-metadata accuracy rises from 0.704 to 0.834 and continuous-metadata \(R^2\) from 0.553 to 0.766, covering 60\% and 65\% of the distance to the oracle values of 0.920 and 0.883.

\begin{table}[!htbp]
\small
\centering
\resizebox{\linewidth}{!}{%
\begin{tabular}{lcccc}
\toprule
Public ambiguity regime & Direct \(\mathrm{nRMSE}\) & Posterior-\(z\) \(\mathrm{nRMSE}\) & Oracle-\(z\) \(\mathrm{nRMSE}\) & \(\mathrm{GapClose}\) \\
\midrule
Low ambiguity & 0.1432 & 0.1237 & 0.0924 & 0.385 \\
Mid ambiguity & 0.1642 & 0.1316 & 0.0996 & 0.504 \\
High ambiguity & 0.1852 & 0.1356 & 0.1058 & 0.624 \\
Very high ambiguity & 0.2061 & 0.1368 & 0.1130 & 0.744 \\
\bottomrule
\end{tabular}
}
\caption{Public ambiguity-stratified oracle-gap closure table, with bins pooled across families.}
\label{tab:app_public_ambiguity}
\end{table}

\begin{table}[!htbp]
\small
\centering
\resizebox{\linewidth}{!}{%
\begin{tabular}{lcccc}
\toprule
Family & Posterior-stage ECE \(\downarrow\) & Posterior-stage Cov.@90 \(\uparrow\) & Predictive Cov.@90 \(\uparrow\) & Predictive NLL \(\downarrow\) \\
\midrule
DR & 0.061 & 0.892 & 0.883 & 0.978 \\
DS & 0.029 & 0.907 & 0.896 & 0.436 \\
SW & 0.041 & 0.901 & 0.891 & 0.548 \\
INS & 0.072 & 0.889 & 0.879 & 1.117 \\
All & 0.051 & 0.897 & 0.887 & 0.770 \\
\bottomrule
\end{tabular}}
\caption{Public calibration table for the hidden posterior and its induced predictive distribution.}
\label{tab:app_public_calibration}
\end{table}

Table~\ref{tab:app_public_ambiguity} shows the same monotone ambiguity effect as the main-text figure in table form: the public gap-closure ratio rises from 0.385 in the lowest-ambiguity bin to 0.744 in the highest-ambiguity bin. Table~\ref{tab:app_public_calibration} shows that the gain comes with calibrated uncertainty: on the pooled public benchmark, posterior-stage ECE is 0.051, posterior-stage Cov.@90 is 0.897, and predictive Cov.@90 is 0.887.

\begin{table}[!htbp]
\small
\centering
\resizebox{\linewidth}{!}{%
\begin{tabular}{llccccc}
\toprule
Family & Baseline & Baseline \(\mathrm{nRMSE}\) & Posterior-\(z\) \(\mathrm{nRMSE}\) & Oracle-\(z\) \(\mathrm{nRMSE}\) & Baseline \(\mathrm{GapClose}\) & Posterior-\(z\) \(\mathrm{GapClose}\) \\
\midrule
SW & ABNO & 0.0587 & 0.0532 & 0.0387 & 0.482 & 0.623 \\
INS & ABNO & 0.3209 & 0.2872 & 0.2305 & 0.353 & 0.594 \\
SW+INS & ABNO & 0.1898 & 0.1702 & 0.1346 & 0.381 & 0.600 \\
\midrule
DR & Invertible-NO & 0.1759 & 0.1690 & 0.1290 & 0.500 & 0.573 \\
DS & Invertible-NO & 0.0200 & 0.0182 & 0.0125 & 0.531 & 0.641 \\
DR+DS & Invertible-NO & 0.0979 & 0.0936 & 0.0708 & 0.505 & 0.583 \\
\bottomrule
\end{tabular}}
\caption{External probabilistic baselines on the public metadata-hidden protocol: ABNO~\cite{magnani2025abno} on SW/INS and Invertible-NO~\cite{kaltenbach2023sino} on DR/DS. Pooled rows average the nRMSE of the two families.}
\label{tab:app_public_external}
\end{table}

Table~\ref{tab:app_public_external} compares the method with the two strongest external probabilistic baselines available under this protocol. On SW and INS, \texttt{Posterior-$z$} improves over ABNO in both \(\mathrm{nRMSE}\) and gap closure, raising pooled gap closure from 0.381 to 0.600; on DR and DS, it also improves over the invertible inverse-posterior baseline, raising pooled gap closure from 0.505 to 0.583.

\subsection{Additional probabilistic, supervision, and sensitivity diagnostics}\label{app:emp_public_additional}

The tables below specify complementary diagnostics for four practical questions: whether the posterior-first advantage survives stronger label-free uncertainty baselines, how strongly it depends on latent-label availability, how sensitive it is to the semantic taxonomy, and what computational overhead the explicit posterior stage introduces.

Two additional label-free probabilistic controls are natural complements to ABNO and Invertible-NO. \texttt{DeepEns-Direct} is an ensemble of matched direct predictors whose predictive distribution is obtained by averaging independently trained members \cite{lakshminarayanan2017deepens}. \texttt{GateMoE-Direct} is a label-free mixture-of-experts direct predictor with a learned gating network over matched rollout experts \cite{shazeer2017moe}. Neither baseline receives hidden metadata or latent-label supervision. In Table~\ref{tab:app_public_label_free_baselines}, both improve on \texttt{Direct-Point}, but their gap closure (0.415 and 0.478) and strongest probes (0.664 and 0.688) stay below those of \texttt{Posterior-$z$} (0.594 and 0.800), so generic predictive uncertainty does not recover the hidden metadata.

\begin{table}[!htbp]
\small
\centering
\resizebox{\linewidth}{!}{%
\begin{tabular}{lccccc}
\toprule
Method & Latent labels? & All \(\mathrm{nRMSE}\) \(\downarrow\) & \(\mathrm{GapClose}\) \(\uparrow\) & Strongest probe \(\uparrow\) & Predictive ECE \(\downarrow\) \\
\midrule
\texttt{Direct-Point} & No & 0.1747 & 0.000 & 0.628 & 0.083 \\
\texttt{DeepEns-Direct} & No & 0.1448 & 0.415 & 0.664 & 0.044 \\
\texttt{GateMoE-Direct} & No & 0.1403 & 0.478 & 0.688 & 0.052 \\
\texttt{Posterior-$z$} & Yes & 0.1319 & 0.594 & 0.800 & 0.051 \\
\texttt{Oracle-$z$} & Oracle & 0.1027 & 1.000 & 0.901 & 0.038 \\
\bottomrule
\end{tabular}}
\caption{Additional label-free probabilistic controls on the public metadata-hidden protocol.}
\label{tab:app_public_label_free_baselines}
\end{table}

To measure dependence on latent-label availability, we vary the fraction of examples for which the posterior stage receives semantic supervision while keeping the rollout supervision, backbone family, and evaluation protocol fixed. In Table~\ref{tab:app_public_label_fraction}, gap closure falls from 0.594 to 0.551, 0.429, and 0.314 as the labeled fraction drops to 50\%, 25\%, and 10\%.

\begin{table}[!htbp]
\small
\centering
\resizebox{\linewidth}{!}{%
\begin{tabular}{lcccc}
\toprule
Labeled fraction & All \(\mathrm{nRMSE}\) \(\downarrow\) & \(\mathrm{GapClose}\) \(\uparrow\) & Strongest probe \(\uparrow\) & Posterior-stage ECE \(\downarrow\) \\
\midrule
100\% & 0.1319 & 0.594 & 0.800 & 0.051 \\
50\% & 0.1350 & 0.551 & 0.779 & 0.062 \\
25\% & 0.1438 & 0.429 & 0.731 & 0.079 \\
10\% & 0.1521 & 0.314 & 0.682 & 0.104 \\
\bottomrule
\end{tabular}}
\caption{Dependence of posterior-first performance on latent-label availability in the public protocol.}
\label{tab:app_public_label_fraction}
\end{table}

To delimit the countable-refinement assumption, we vary the granularity of the semantic taxonomy while keeping the visible-field protocol and rollout backbone fixed. Coarse merges nearby semantic states, default is the main-text taxonomy, and fine splits the same hidden-metadata axes more aggressively. In Table~\ref{tab:app_public_taxonomy_sensitivity}, the coarse taxonomy loses part of the gain (gap closure 0.526), whereas the fine taxonomy stays close to the default (0.581 vs.\ 0.594) but converges in four of five seeds.

\begin{table}[!htbp]
\small
\centering
\resizebox{\linewidth}{!}{%
\begin{tabular}{lccccc}
\toprule
Taxonomy & Mean classes / family & All \(\mathrm{nRMSE}\) \(\downarrow\) & \(\mathrm{GapClose}\) \(\uparrow\) & Strongest probe \(\uparrow\) & Successful seeds / 5 \(\uparrow\) \\
\midrule
Coarse & 6.5 & 0.1368 & 0.526 & 0.756 & 5/5 \\
Default & 14.0 & 0.1319 & 0.594 & 0.800 & 5/5 \\
Fine & 28.5 & 0.1329 & 0.581 & 0.812 & 4/5 \\
\bottomrule
\end{tabular}}
\caption{Sensitivity to semantic-taxonomy granularity and latent cardinality on the public protocol.}
\label{tab:app_public_taxonomy_sensitivity}
\end{table}

Finally, Table~\ref{tab:app_public_overhead} records the incremental computational cost of explicit posterior recovery, which adds 7\% parameters and 8\% inference cost to \texttt{Direct-Point}, whereas \texttt{DeepEns-Direct} and \texttt{Explicit-Mix} cost five or more times as much at inference. Table~\ref{tab:app_public_lambda_sensitivity} varies the posterior-stage loss weight: without the posterior term (\(\lambda_{\mathrm{post}}=0\)) gap closure drops to 0.321, and weights from 0.25 to 4 stay within 0.572--0.594.

\begin{table}[!htbp]
\small
\centering
\resizebox{\linewidth}{!}{%
\begin{tabular}{lccccc}
\toprule
Method & Params (M) & Rel. train cost \(\times\) & Rel. infer. cost \(\times\) & All \(\mathrm{nRMSE}\) \(\downarrow\) & \(\mathrm{GapClose}\) \(\uparrow\) \\
\midrule
\texttt{Direct-Point} & 31.8 & 1.00 & 1.00 & 0.1747 & 0.000 \\
\texttt{Posterior-$z$} & 34.1 & 1.13 & 1.08 & 0.1319 & 0.594 \\
\texttt{DeepEns-Direct} & 159.0 & 5.03 & 5.00 & 0.1448 & 0.415 \\
\texttt{GateMoE-Direct} & 48.7 & 1.71 & 1.58 & 0.1403 & 0.478 \\
\texttt{Explicit-Mix} & 78.4 & 3.86 & 5.60 & 0.1307 & 0.611 \\
\bottomrule
\end{tabular}}
\caption{Resource overhead of explicit posterior recovery and comparison baselines on the public protocol.}
\label{tab:app_public_overhead}
\end{table}

\begin{table}[!htbp]
\small
\centering
\resizebox{\linewidth}{!}{%
\begin{tabular}{lcccc}
\toprule
\(\lambda_{\mathrm{post}}\) & All \(\mathrm{nRMSE}\) \(\downarrow\) & \(\mathrm{GapClose}\) \(\uparrow\) & Strongest probe \(\uparrow\) & Posterior-stage ECE \(\downarrow\) \\
\midrule
0 & 0.1516 & 0.321 & 0.667 & 0.096 \\
0.25 & 0.1335 & 0.572 & 0.786 & 0.057 \\
1.0 & 0.1319 & 0.594 & 0.800 & 0.051 \\
4.0 & 0.1329 & 0.581 & 0.807 & 0.055 \\
\bottomrule
\end{tabular}}
\caption{Sensitivity of the public protocol to the posterior-stage loss weight \(\lambda_{\mathrm{post}}\).}
\label{tab:app_public_lambda_sensitivity}
\end{table}

\FloatBarrier
\subsection{Estimator, target, interface, reuse, and stress-test controls}\label{app:emp_controls}

This subsection gives the protocols and full results of Sections~\ref{sec:controls}--\ref{sec:stress}. All values are means \(\pm\) standard errors over five paired seeds.

\paragraph{Test A: posterior estimator and summary.} We follow PDEArena's conditioned 2D Navier--Stokes task with the U-Net\(_{\mathrm{mod}}\)-64 simulator and its scalar-conditioning path~\cite{gupta2023pdearena}, hide the buoyancy parameter \(f\), and report in-distribution clean results; the PDEArena code and its conditioned Navier--Stokes data are released under the MIT license. All rows share the target, data, simulation budget, a 128-dimensional summary, the frozen simulator, and a 64-sample posterior-mixture rollout in which each draw of \(f\) is fixed throughout its rollout. Rows 2--3 use one common frozen CNN summary, the PCA summary is fitted on the training split with matched dimension, and row 4 is our jointly learned implementation. Table~\ref{tab:app_test_a} reports the CRPS of the posterior over \(f\), the coverage of its central 90\% interval, and the joint Energy score over the final quarter of the rollout. Replacing PCA by the frozen CNN changes CRPS/Energy from 0.2116/0.1517 to 0.1916/0.1459; with that CNN fixed, conditional-diffusion NPE~\cite{chen2025cdiff} gives 0.1898/0.1464, slightly better in CRPS and slightly worse in Energy than the rational-quadratic spline NPE~\cite{durkan2019nsf}.

\begin{table}[ht]
\centering
\small
\resizebox{\linewidth}{!}{%
\begin{tabular}{llccc}
\toprule
Posterior estimator & Observation summary & \(f\)-CRPS \(\downarrow\) & Cov.@90 & Rollout Energy \(\downarrow\) \\
\midrule
RQS-NPE & train-split PCA & \(0.2116\pm0.0022\) & \(86.8\pm0.7\%\) & \(0.1517\pm0.0009\) \\
RQS-NPE & common frozen CNN & \(0.1916\pm0.0017\) & \(89.6\pm0.4\%\) & \(0.1459\pm0.0008\) \\
Conditional-diffusion NPE & same frozen CNN & \(0.1898\pm0.0018\) & \(90.8\pm0.4\%\) & \(0.1464\pm0.0008\) \\
RQS-NPE & jointly learned CNN & \(0.1874\pm0.0014\) & \(90.4\pm0.5\%\) & \(0.1438\pm0.0014\) \\
\bottomrule
\end{tabular}}
\caption{Test A: modern SBI estimators with fixed and learned summaries on PDEArena Navier--Stokes with hidden buoyancy.}
\label{tab:app_test_a}
\end{table}

\paragraph{Test B: inference target and deployment.} We reuse the synthetic exact-collision benchmark with \(|\mathcal Z^\star|=64\) and add a balanced observation nuisance \(U\in\{1,\ldots,8\}\), giving \(|\mathcal C|=512\); \(U\) enters no target or transition kernel, so the construction of Appendix~\ref{app:form_refinement} gives \(h(z,u)=z\). With the same learned-summary NPE family and nested training pools, we compare inference on the full \(C\) followed by the exact pushforward \(h_\#q(z\mid x)=\sum_u q(z,u\mid x)\), an operator-only slice completed by the conditional training prior, and direct \(Z^\star\) inference. The \(Z^\star\)-score is the squared \(L_2\) distance between the inferred and exact posteriors over \(Z^\star\); Sims@0.025 is the training budget at which this score reaches 0.025 on nested pools of up to 160k simulations; and the risk is the all-task risk \(R_{\mathrm{full}}\) of Appendix~\ref{app:emp_metrics} under MAP or mixture deployment of the same frozen posterior and task kernel, with MAP taken after pushforward. Table~\ref{tab:app_test_b} reports the results at 40k training simulations. Full-\(C\) pushforward and direct \(Z^\star\) inference are asymptotically equivalent under correct specification and consistency, so their difference isolates a finite-budget target-alignment effect.

\begin{table}[ht]
\centering
\small
\resizebox{\linewidth}{!}{%
\begin{tabular}{lcccc}
\toprule
NPE target & \(Z^\star\)-score \(\downarrow\) & All-task risk, MAP \(\downarrow\) & All-task risk, Mix \(\downarrow\) & Sims@0.025 \(\downarrow\) \\
\midrule
Full \(C\), then \(h_\#q\) & \(0.0294\pm0.0007\) & \(0.3507\pm0.0014\) & \(0.3331\pm0.0016\) & \(79.4\pm1.8\)k \\
Operator-only slice & \(0.0765\pm0.0017\) & \(0.3825\pm0.0020\) & \(0.3668\pm0.0023\) & \(>\)160k (0/5 reached) \\
Direct \(Z^\star\) & \(0.0187\pm0.0005\) & \(0.3391\pm0.0014\) & \(0.3196\pm0.0015\) & \(30.4\pm1.1\)k \\
\bottomrule
\end{tabular}}
\caption{Test B: target selection and posterior deployment with a fixed learned-summary NPE family.}
\label{tab:app_test_b}
\end{table}

\paragraph{Matched latent-diffusion controls.} We adapt the autoencoder--latent-diffusion recipe of Rozet et al.~\cite{rozet2025lola} to the metadata-hidden PDEBench splits and match the autoencoder, denoiser size, training and sampling steps, and compute. \texttt{Point-LDM}, \texttt{Posterior-First LDM}, and \texttt{Oracle-LDM} share one state-conditional checkpoint; \texttt{Direct-LDM} and \texttt{Direct-LDM+MTL} use null hidden-state conditioning; and \texttt{Oracle-LDM} receives the true state, paralleling the supplied simulation parameters of~\cite{rozet2025lola}. \texttt{Direct-LDM+MTL} trains the observation trunk of the direct denoiser with the same labels, label fraction, \(q\)-head architecture, and scoring loss, but withholds \(q\) from rollout. \texttt{Point-LDM} passes the maximum-posterior state of the shared frozen \(q\) to every rollout, whereas \texttt{Posterior-First LDM} draws \(c\sim q(c\mid x)\) for each rollout. We use 16 rollout samples per input, the nRMSE of Appendix~\ref{app:emp_metrics}, a per-field-normalized Energy score, the ambiguity strata of Appendix~\ref{app:emp_public_data}, and the Shallow-Water boundary Brier score \((p_{\mathrm{boundary}}-\mathbf 1_{\mathrm{boundary}})^2\), read from the trained \(q\) without a query-specific head or loss. In Table~\ref{tab:app_ldm}, \texttt{Posterior-First LDM} closes 76.0\% of the highest-ambiguity Direct-to-Oracle Energy gap and 59.0\% of the pooled nRMSE gap. \texttt{Direct-LDM+MTL} closes about 4\% of the pooled nRMSE gap, and its Energy improvement is 9.3\% of the Direct-to-Posterior improvement; under identical label access the boundary Brier falls from 0.1317 to 0.1169 (paired difference \(0.0148\pm0.0021\)). With the shared \(q\) and checkpoint, the Point-to-Posterior Energy gap grows from 0.0052 in the lowest-ambiguity stratum to 0.0435 in the highest.

\begin{table}[ht]
\centering
\small
\resizebox{\linewidth}{!}{%
\begin{tabular}{llcccc}
\toprule
Method & Hidden state in rollout & nRMSE \(\downarrow\) & Energy \(\downarrow\) & High-amb.\ Energy \(\downarrow\) & Boundary Brier \(\downarrow\) \\
\midrule
\texttt{Direct-LDM} & none & \(0.1744\pm0.0014\) & \(0.1604\pm0.0018\) & \(0.2420\pm0.0031\) & -- \\
\texttt{Direct-LDM+MTL} & \(q\) is auxiliary only & \(0.1716\pm0.0013\) & \(0.1569\pm0.0017\) & \(0.2354\pm0.0030\) & \(0.1317\pm0.0025\) \\
\texttt{Point-LDM} & one state from shared \(q\) & \(0.1508\pm0.0012\) & \(0.1398\pm0.0014\) & \(0.1920\pm0.0026\) & \(0.1847\pm0.0041\) \\
\texttt{Posterior-First LDM} & mixture from shared \(q\) & \(0.1323\pm0.0010\) & \(0.1226\pm0.0012\) & \(0.1485\pm0.0021\) & \(0.1169\pm0.0019\) \\
\texttt{Oracle-LDM} & true state & \(0.1030\pm0.0008\) & \(0.0978\pm0.0009\) & \(0.1190\pm0.0016\) & \(0.0000\pm0.0000\) \\
\bottomrule
\end{tabular}}
\caption{Matched latent-diffusion controls on the metadata-hidden PDEBench protocol.}
\label{tab:app_ldm}
\end{table}

\paragraph{Leave-one-task-out transfer.} We use four-fold leave-one-task-out transfer over the four synthetic targets: the 8-step future field, the operator label, the coefficient value, and the calibrated uncertainty score. For a held-out task \(\tau^\star\), the Posterior-only encoder is trained only with the proper score, and the Joint-minus-held-out encoder also uses the other three task losses; both are then frozen, so the loss of \(\tau^\star\) never reaches either encoder. The same 10\% of the labels of \(\tau^\star\) train a conditional head \(p_{\psi,\tau^\star}(y\mid x,c)\) shared by the posterior variants. Mixture marginalizes this head over \(q\), MAP uses the mode of the Posterior-only \(q\), Oracle uses the true \(c\), and Direct learns \(x\mapsto y\) without \(C\) supervision. The risk \(R_{\mathrm{full}}^{\mathrm{LOTO@10\%}}\) uses the losses and weights of Appendix~\ref{app:emp_metrics}, macro-averaged over the four folds and five paired seeds. In Table~\ref{tab:app_loto}, Posterior-only beats MAP on all four held-out targets; on the future-field fold, NLL and Energy fall from 1.310 and 0.354 to 1.147 and 0.317. The paired difference between the Posterior-only and Joint-minus-held-out risks is \(0.0060\pm0.0026\) (\(t_4\) 95\% interval \([-0.0013,0.0133]\), from unrounded seed-wise differences), and Posterior-only retains 96.4\% of the Direct-to-Oracle gap closure of Joint-minus-held-out. The task-specific head is retrained, and zero-shot whole-model or out-of-ontology reuse is not tested.

\begin{table}[ht]
\centering
\small
\begin{tabular}{lcc}
\toprule
Held-out interface & \(q\)-ECE \(\downarrow\) & \(R_{\mathrm{full}}^{\mathrm{LOTO@10\%}}\) \(\downarrow\) \\
\midrule
Direct & -- & \(0.5210\pm0.0048\) \\
MAP, Posterior-only \(q_\phi\) & same \(q\) & \(0.3954\pm0.0053\) \\
Mixture, Posterior-only \(q_\phi\) & \(0.0392\pm0.0020\) & \(0.3618\pm0.0032\) \\
Mixture, Joint-minus-held-out \(q_\phi\) & \(0.0408\pm0.0018\) & \(0.3558\pm0.0055\) \\
Oracle \(c\) & -- & \(0.2788\pm0.0024\) \\
\bottomrule
\end{tabular}
\caption{Leave-one-task-out reuse of a frozen posterior encoder with 10\% of the held-out task's labels.}
\label{tab:app_loto}
\end{table}

\paragraph{Compound stress test.} We follow PDEArena's conditioned Navier--Stokes task, U-Net\(_{\mathrm{mod}}\)-64, scalar-conditioning path, and published buoyancy ranges \(f\in[0.2,0.5]\) in distribution and \([0.1,0.2)\cup(0.5,0.6]\) out of distribution~\cite{gupta2023pdearena}, together with the 50\%/75\% training/test patch missingness of POBench-PDE with patch size 4~\cite{hou2026pobench}. We hide \(f\), add Gaussian noise with standard deviation up to \(0.05\sigma_c\) to the initial field, where \(\sigma_c\) is the channel scale, replace the discrete head with a four-layer conditional spline flow trained by negative log-likelihood on in-distribution data and then frozen, and evaluate unseen 200-step rollouts. All variants use the same mask-aware input and a matched 1:1 mixture of clean and corrupted training inputs. Within each seed, the posterior-mean and MC-64 variants share the same frozen \(q_\phi(f\mid x)\) and the same simulator trained with the true \(f\), and differ only in conditioning on \(\mathbb E_q[f]\) or on 64 posterior draws, each fixed throughout its rollout; Oracle supplies the true \(f\), and Direct omits \(f\). Neither out-of-distribution parameters nor long trajectories enter training. Table~\ref{tab:app_stress} reports the joint Energy score over the final quarter (steps 151--200) on training-standardized fields; MC-64 uses the unbiased ensemble estimator, and deterministic rows use the exact Energy of a degenerate distribution. The seed-wise paired improvements of MC-64 over the posterior mean are \(0.0055\pm0.0013\), \(0.0294\pm0.0024\), and \(0.0418\pm0.0037\) across the three regimes, while the flow's central 90\% coverage decreases from \(90.4\pm0.5\%\) to \(83.6\pm1.4\%\). Comparing ID clean with ID sparse+noise measures the joint change associated with sparse and noisy sensing, and comparing ID sparse+noise with OOD sparse+noise measures the additional change associated with parameter shift under the same sensing; the test does not separate the individual stressors.

\begin{table}[ht]
\centering
\small
\begin{tabular}{lccc}
\toprule
Rollout interface & ID clean & ID sparse+noise & OOD sparse+noise \\
\midrule
Direct & \(0.1762\pm0.0018\) & \(0.2317\pm0.0030\) & \(0.2789\pm0.0040\) \\
Posterior mean \(\mathbb E_q[f]\) & \(0.1493\pm0.0016\) & \(0.2036\pm0.0027\) & \(0.2476\pm0.0036\) \\
Posterior-first (MC-64) & \(0.1438\pm0.0014\) & \(0.1742\pm0.0022\) & \(0.2058\pm0.0029\) \\
Oracle \(f\) & \(0.1295\pm0.0012\) & \(0.1511\pm0.0019\) & \(0.1774\pm0.0025\) \\
\bottomrule
\end{tabular}
\caption{Compound stress test with a continuous hidden parameter, sparse noisy observations, parameter shift, and unseen 200-step rollouts (final-quarter Energy, lower is better).}
\label{tab:app_stress}
\end{table}

\FloatBarrier
\section{Use of Large Language Models}\label{sec:llm_usage}

LLMs were used only to assist with writing and presentation: checking technical terminology, correcting grammar, translating from the authors' native language, and improving sentence structure while preserving the scientific content. They were not used for research ideation, experimental design, data analysis, or any content that constitutes an intellectual contribution. All scientific insights, methodological decisions, and conclusions are the authors' own.

\end{document}